\documentclass[format=acmsmall, review=false, screen=true]{acmart}

\usepackage{booktabs} 

\usepackage[ruled]{algorithm2e} 

\SetAlFnt{\small}
\SetAlCapFnt{\small}
\SetAlCapNameFnt{\small}
\SetAlCapHSkip{0pt}
\IncMargin{-\parindent}

\usepackage{booktabs} 
\usepackage{epstopdf}
\usepackage[ruled]{algorithm2e} 
\usepackage{algorithmic}
\usepackage{graphicx}
\usepackage{subfigure}
\usepackage{placeins}
\usepackage{url}
\usepackage{amssymb}
\usepackage{amsmath}
\usepackage{graphicx}
\usepackage{epstopdf}
\usepackage{rotating}
\usepackage{array,multirow}

\def \f {\mathbf{f}}

\def \x {\mathbf{x}}
\def \u {\mathbf{u}}

\def \u {\mathbf{u}}

\def \w {\mathbf{w}}

\def \D {\mathcal{D}}

\def \I {\mathbb{I}}
\def \R {\mathbb{R}}

\def \N {\mathcal{N}}

\def \bq {\begin{eqnarray}}
\def \eq {\end{eqnarray}}
\def \bqs {\begin{eqnarray*}}
	\def \eqs {\end{eqnarray*}}
\def \sign {\textrm{sign}}

\begin{document}
\title[Malicious URL Detection using Machine Learning: A Survey]{Malicious URL Detection using Machine Learning: A Survey}

\author{Doyen Sahoo}
\affiliation{%
	\institution{\\School of Information Systems, Singapore Management University}}
\author{Chenghao Liu}
\affiliation{%
	\institution{\\School of Information Systems, Singapore Management University}}
\author{Steven C.H. Hoi}
\affiliation{%
	\institution{\\School of Information Systems, Singapore Management University \\ Salesforce Research Asia}}
\email{{doyens,chliu,chhoi}@smu.edu.sg}

\begin{abstract}
Malicious URL, a.k.a. malicious website, is a common and serious threat to cybersecurity. Malicious URLs host unsolicited content (spam, phishing, drive-by downloads, etc.) and lure unsuspecting users to become victims of scams (monetary loss, theft of private information, and malware installation), and cause losses of billions of dollars every year. It is imperative to detect and act on such threats in a timely manner. Traditionally, this detection is done mostly through the usage of blacklists. However, blacklists cannot be exhaustive, and lack the ability to detect newly generated malicious URLs. To improve the generality of malicious URL detectors, machine learning techniques have been explored with increasing attention in recent years. This article aims to provide a comprehensive survey and a structural understanding of Malicious URL Detection techniques using machine learning. We present the formal formulation of Malicious URL Detection as a machine learning task, and categorize and review the contributions of literature studies that addresses different dimensions of this problem (feature representation, algorithm design, etc.). Further, this article provides a timely and comprehensive survey for a range of different audiences, not only for machine learning researchers and engineers in academia, but also for  professionals and practitioners in cybersecurity industry, to help them understand the state of the art and facilitate their own research and practical applications. We also discuss practical issues in system design, open research challenges, and point out important directions for future research.
\end{abstract}

%
%
%

\maketitle
\renewcommand{\shortauthors}{Sahoo, Liu and Hoi}

\section{Introduction}


The advent of new communication technologies has had tremendous impact in the growth and promotion of businesses spanning across many applications including online-banking, e-commerce, and social networking. In fact, in today's age it is almost mandatory to have an online presence to run a successful venture. As a result, the importance of the World Wide Web has continuously been increasing. Unfortunately, the technological advancements come coupled with new sophisticated techniques to attack and scam users. Such attacks include rogue websites that sell counterfeit goods, financial fraud by tricking users into revealing sensitive information which eventually lead to theft of money or identity, or even installing malware in the user's system. There are a wide variety of techniques to implement such attacks, such as explicit hacking attempts, drive-by download, social engineering, phishing, watering hole, man-in-the middle, SQL injections, loss/theft of devices, denial of service, distributed denial of service, and many others. Considering the variety of attacks, potentially new attack types, and the innumerable contexts in which such attacks can appear, it is hard to design robust systems to detect cyber-security breaches. The limitations of traditional security management technologies are becoming more and more serious given this exponential growth of new security threats, rapid changes of new IT technologies, and significant shortage of security professionals. Most of these attacking techniques are realized through spreading compromised URLs (or the spreading of such URLs forms a critical part of the attacking operation) \cite{hong2012state}.

\begin{figure*}[ht!]
	\centering
	\includegraphics[width=100mm]{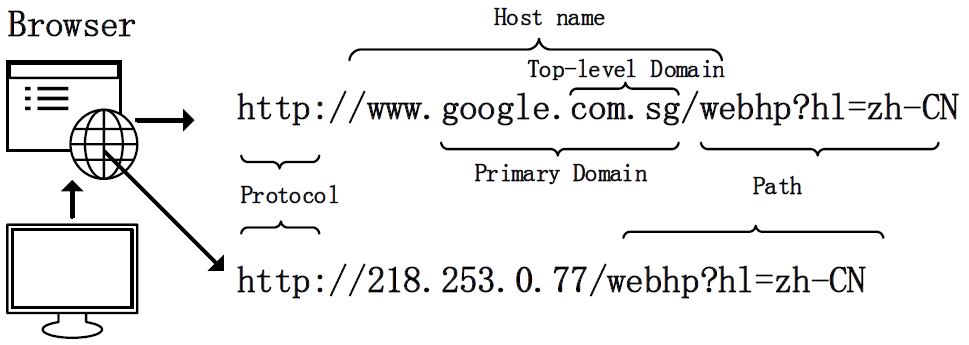}
	\caption{Example of a URL - ``Uniform Resource Locator"}
	\label{fig:url}\vspace{-0.1in}
\end{figure*}

URL is the abbreviation of Uniform Resource Locator, which is the global address of documents and other resources on the World Wide Web. A URL has two main components : (i) protocol identifier (indicates what protocol to use) (ii) resource name (specifies the IP address or the domain name where the resource is located). The protocol identifier and the resource name are separated by a colon and two forward slashes, e.g. Figure \ref{fig:url}.

Compromised URLs that are used for cyber attacks are termed as \emph{malicious URLs}. In fact, it was noted that close to one-third of all websites are potentially malicious in nature \cite{liang2009malicious}, demonstrating rampant use of malicious URLs to perpetrate cyber-crimes.  A Malicious URL or a malicious web site hosts a variety of unsolicited content in the form of spam, phishing, or drive-by download in order to launch attacks. Unsuspecting users visit such web sites and become victims of various types of scams, including monetary loss, theft of private information (identity, credit-cards, etc.), and malware installation. Popular types of attacks using malicious URLs include: Drive-by Download, Phishing and Social Engineering, and Spam \cite{patil2015survey}. Drive-by download \cite{cova2010detection} refers to the (unintentional) download of malware upon just visiting a URL. Such attacks are usually carried out by exploiting vulnerabilities in plugins or inserting malicious code through JavaScript. Phishing and Social Engineering attacks \cite{heartfield2015taxonomy} trick the users into revealing private or sensitive information by pretending to be genuine web pages. Spam is the usage of unsolicited messages for the purpose of advertising or phishing. These attacks occur in large numbers and have caused billions of dollars worth of damage, some even exploiting natural disasters \cite{verma2018phishing}. Effective systems to detect such malicious URLs in a timely manner can greatly help to counter large number of and a variety of cyber-security threats. Consequently, researchers and practitioners have worked to design effective solutions for Malicious URL Detection.

The most common method to detect malicious URLs deployed by many antivirus groups is the blacklist method. Blacklists are essentially a database of URLs that have been confirmed to be malicious in the past. This database is compiled over time (often through crowd-sourcing solutions, e.g. PhishTank \cite{opendns2016phishtank}), as and when it becomes known that a URL is malicious. Such a technique is extremely fast due to a simple query overhead, and hence is very easy to implement. Additionally, such a technique would (intuitively) have a very low false-positive rate (although, it was reported that often blacklisting suffered from non-trivial false-positive rates \cite{sinha2008shades}). However, it is almost impossible to maintain an exhaustive list of malicious URLs, especially since new URLs are generated everyday. Attackers use creative techniques to evade blacklists and fool users by modifying the URL to ``appear" legitimate via obfuscation. Garera et. al. \cite{garera2007framework} identified four types of obfuscation: Obfuscating the Host with an IP, Obfuscating the Host with another domain, Obfuscating the host with large host names, and misspelling. All of these try to hide the malicious intentions of the website by masking the malicious URL. Recently, with the increasing popularity of URL shortening services, it has become a new and widespread obfuscation technique (hiding the malicious URL behind a short URL) \cite{chhabra2011phi,alshboul2015detecting}. Once the URLs appear legitimate, users visit them, and an attack can be launched. This is often done by malicious code embedded into the JavaScript. Often attackers will try to obfuscate the code so as to prevent signature based tools from detecting them. Attackers use many other techniques to evade blacklists including: fast-flux, in which proxies are automatically generated to host the webpage; algorithmic generation \cite{li2019domain} of new URLs; etc. Additionally, attackers can simultaneously launch more than one attack to alter the attack-signature, making it undetectable by tools that focus on specific signatures. Blacklisting methods, thus have severe limitations, and it appears almost trivial to bypass them, especially because  blacklists are useless for making predictions on new URLs.

To overcome these issues, in the last decade, researchers have applied machine learning techniques for Malicious URL Detection \cite{garera2007framework,mcgrath2008behind,ma2009beyond,ma2011learning,purkait2012phishing,khonji2013phishing,patil2015survey,nepali2016you,kuyama2016method}. Machine Learning approaches, use a set of URLs as training data, and based on the statistical properties, learn a prediction function to classify a URL as malicious or benign. This gives them the ability to generalize to new URLs unlike blacklisting methods. The primary requirement for training a machine learning model is the presence of training data. In the context of malicious URL detection, this would correspond to a set of large number of URLs. Machine learning can broadly be classified into supervised, unsupervised, and semi-supervised, which correspond to having the labels for the training data, not having the labels, and having labels for limited fraction of training data, respectively. Labels correspond to the knowledge that a URL is malicious or benign.

After the training data is collected, the next step is to extract informative features such that they sufficiently describe the URL and at the same time, they can be interpreted mathematically by machine learning models. For example, simply using the URL string directly may not allow us to learn a good prediction model (which in some extreme cases may reduce the prediction model to a blacklist method).
Thus, one would need to extract suitable features based on some principles or heuristics to obtain a good feature representation of the URL. This may include lexical features (statistical properties of the URL string, bag of words, n-gram, etc.), host-based features (WHOIS info, geo-location properties of the host, etc.), etc. These features and other features that are used in for this task will be discussed in much greater detail in this survey. These features after being extracted have to be processed into a suitable format (e.g. a numerical vector), such that they can be plugged into an off-the-shelf machine learning method for model training. The ability of these features to provide relevant information is critical to subsequent machine learning, as the underlying assumption of machine learning (classification) models is that feature representations of the malicious and benign URLs have different distributions. Therefore, the quality of feature representation of the URLs is critical to the quality of the resulting malicious URL predictive model learned by machine learning.

Using the training data with the appropriate feature representation, the next step in building the prediction model is the actual training of the model. There are plenty of classification algorithms can be directly used over the training data (Naive Bayes, Support Vector Machine, Logistic Regression, etc.). However, there are certain properties of the URL data that may make the training difficult (both in terms of scalability and learning the appropriate concept). For example, the number of URLs available for training can be in the order of millions (or even billions). As a result, the training time for traditional models may be too high to be practical. Consequently, Online Learning \cite{hoi2014libol,hoi2018online}, a family of scalable learning techniques have been heavily applied for this task. Another challenge is the sparsity of the bag-of-words (BoW) feature representation of the URLs. 
These features indicate whether a particular word (or string) appears in a URL or not - as a result every possible type of word that may appear in any URL becomes a feature. This representation may result in millions of features which would be very sparse (most features are absent most of the time, as a URL will usually have very few of the millions of possible words present in it). Accordingly, a learning method should exploit this sparsity to improve learning efficiency and efficacy. There are other challenges which are specifically found for this task, and have warranted appropriate contributions in machine learning methodology to alleviate these challenges. 

In this survey, we review the state-of-the-art machine learning techniques for malicious URL detection in literature. We specifically focus on the contributions made for feature representation and learning algorithm development in this domain. We systematically categorize the various types of feature representation used for creating the training data for this task, and also categorize various learning algorithms used to learn a good prediction model. We also discuss the open research problems and identify directions for future research. We first discuss the broad categories of strategies used for detecting malicious URLs - Blacklists (and Heuristics) and Machine Learning. We formalize the setting as a machine learning problem, where the primary requirement is good feature representation and the learning algorithm used. We then comprehensively present various types of feature representation used for this problem. This is followed by presenting various algorithms that have been used to solve this task, and have been developed based on the properties of URL data. Finally we discuss the newly emerging concept of Malicious URL Detection as a service and the principles to be used while designing such a system. We end the survey by discussing the practical issues and open problems.

\begin{figure*}[ht!]
	\vspace{-0.1in}
	\centering
	\includegraphics[width=135mm]{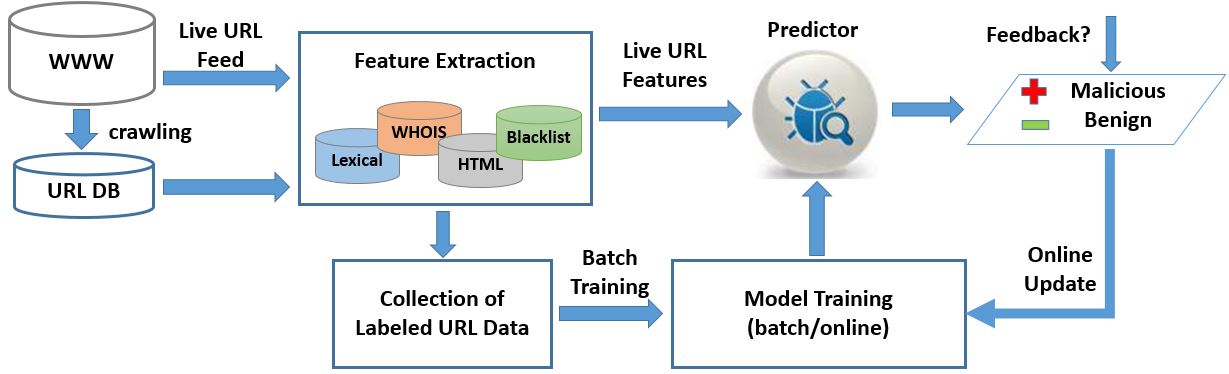}
	\caption{A general processing framework for Malicious URL Detection using Machine Learning}
	\label{fig:framework}\vspace{-0.1in}
\end{figure*}

\section{Malicious URL Detection}

We first present the key principles to solve Malicious URL detection, followed by formalizing it as a machine learning task.

\subsection{Overview of Principles of Detecting Malicious URLs }

A variety of approaches have been attempted to tackle the problem of Malicious URL Detection. According to the fundamental principles, we categorize them into: (i) Blacklisting or  Heuristics, and (ii) Machine Learning approaches \cite{canali2011prophiler,eshete2013binspect}.

\subsubsection{Blacklisting or Heuristic Approaches}
\emph{Blacklisting} approaches are a common and classical technique for detecting malicious URLs, which often maintain a list of URLs that are known to be malicious. Whenever a new URL is visited, a database lookup is performed. If the URL is present in the blacklist, it is considered to be malicious and then a warning will be generated; else it is assumed to be benign. Blacklisting suffers from the inability to maintain an exhaustive list of all possible malicious URLs, as new URLs can be easily generated daily, thus making it impossible for them to detect new threats \cite{sheng2009empirical}. This is particularly of critical concern when the attackers generate new URLs algorithmically, and can thus bypass all blacklists. Despite several problems faced by blacklisting \cite{sinha2008shades}, due to their simplicity and efficiency, they continue to be one of the most commonly used techniques by many anti-virus systems today.

\emph{Heuristic} approaches are a kind of extension of Blacklist methods, wherein the idea is to create a ``blacklist of signatures". Common attacks are identified, and a signature is assigned to this attack type. Intrusion Detection Systems can scan the web pages for such signatures, and raise a flag if some suspicious behavior is found. These methods have better generalization capabilities than blacklisting, as they have the ability to detect threats in new URLs as well. However, such methods can be designed for only a limited number of common threats, and can not generalize to all types of (novel) attacks. Moreover, using obfuscation techniques, it is not difficult to bypass them. A more specific version of heuristic approaches is through analysis of execution dynamics of the webpage (e.g. \cite{moshchuk2007spyproxy,rieck2010cujo,qassrawi2011detecting,kim2011suspicious,kolbitsch2012rozzle,shibahara2017malicious} etc.). Here also, the idea is to look for a signature of malicious activity such as unusual process creation, repeated redirection, etc. These methods necessarily require visiting the webpage and thus the URLs actually can make an attack. As a result, such techniques are often implemented in controlled environment like a disposable virtual machine. Such techniques are very resource intensive, and require all execution of the code (including the rich client sided code). Another drawback is that websites may not launch an attack immediately after being visited, and thus may go undetected.

\subsubsection{Machine Learning Approaches}
These approaches try to analyze the information of a URL and its corresponding websites or webpages, by extracting good feature representations of URLs, and training a prediction model on training data of both malicious and benign URLs. There are two-types of features that can be used - \emph{static} features, and \emph{dynamic} features. In static analysis, we perform the analysis of a webpage based on information available without executing the URL (i.e., executing JavaScript, or other code) \cite{ma2011learning,ma2009beyond,ma2009identifying,eshete2013binspect}. The features extracted include lexical features from the URL string, information about the host, and sometimes even HTML and JavaScript content. Since no execution is required, these methods are safer than the Dynamic approaches. The underlying assumption is that the distribution of these features is different for malicious and benign URLs. Using this distribution information, a prediction model can be built, which can make predictions on new URLs. Due to the relatively safer environment for extracting important information, and the ability to generalize to all types of threats (not just common ones which have to be defined by a signature), static analysis techniques have been extensively explored by applying machine learning techniques. In this survey, we focus on static analysis techniques where machine learning has found tremendous success. Dynamic analysis techniques include monitoring the behavior of the systems which are potential victims, to look for any anomaly. These include \cite{canfora2014detection} which monitor the system call sequences for abnormal behavior, and \cite{tao2014suspicious} which mine internet access log data for suspicious activity. Dynamic analysis techniques have inherent risks, and are difficult to implement and generalize.

Next, we formalize the problem of malicious URL detection as a machine learning task (specifically binary classification) which allows us to generalize most of the existing work in literature. Alternate problem settings will also be discussed in Section \ref{sec:ml}.

\subsection{Problem Formulation}
We formulate Malicious URL detection as a binary classification task for two-class prediction: ``malicious" versus ``benign". Specifically, given a data set with $T$ URLs $\{(\u_1, y_1), \dots, (\u_T, y_T)\}$, where $\u_t $ for $t = 1,\dots, T$ represents a URL from the training data, and $y_t \in \{1, -1\}$ is the corresponding label where $y_t = 1$ represents a malicious URL and $y_t=-1$ represents a benign URL. The crux to automated malicious URL detection is two-fold:
\begin{enumerate}
	\item \emph{Feature Representation}: Extracting the appropriate feature representation: $\u_t \rightarrow \x_t$ where $\x_t \in \R^d$ is a $d$-dimensional feature vector representing the URL; and
	\item \emph{Machine Learning}: Learning a prediction function $\f: \R^d \rightarrow \R$ which predicts the class assignment for any URL instance $\x$ using proper feature representations.
\end{enumerate}

The goal of machine learning for malicious URL detection is to maximize the predictive accuracy. Both of the folds above are important to achieve this goal. While the first part of feature representation is often based on domain knowledge and heuristics, the second part focuses on training the classification model via a data driven optimization approach. Fig. \ref{fig:framework} illustrates a general work-flow for Malicious URL Detection using machine learning.

The first key step is to convert a URL $\u$ into a feature vector $\x$, where several types of information can be considered and different techniques can be used. Unlike learning the prediction model, this part cannot be directly computed by a mathematical function (not for most of it). Using domain knowledge and related expertise, a feature representation is constructed by crawling all relevant information about the URL. These range from lexical information (length of URL, the words used in the URL, etc.) to host-based information (WHOIS info, IP address, location, etc.). Once the information is gathered, it is processed to be stored in a feature vector $\x$. Numerical features can be stored in $\x$ as is, and identity related information or lexical features are usually stored through a binarization or bag-of-words (BoW) approach. Based on the type of information used, $\x \in \R^d$ generated from a URL is a $d$-dimensional vector where $d$ can be less than $100$ or can be in the order of millions. A unique challenge that affects this problem setting is that the number of features may not be fixed or known in advance. For example, using a BoW approach one can track the occurrence for every type of word that may have occurred in a URL in the training data. A model can be trained on this data, but while predicting, new URLs may have words that did not occur in the training data. It is thus a challenging task to design a good feature representation that is robust to unseen data.

After obtaining the feature vector $\x$ for the training data, to learn the prediction function $\f: \R^d \rightarrow \R$, it is usually formulated as an optimization problem such that the detection accuracy is maximized (or alternately, a loss function is minimized). The function $\f$ is (usually) parameterized by a $d-$ dimensional weight vector $\w$, such that $\f(\x) = (\w^\top\x)$. Let $\hat{y_t} = \sign (\f(\x_t))$ denote the class label prediction made by the function $\f$. The number of mistakes made by the prediction model on the entire training data is given by: $\sum_{t=1}^T\I_{\hat{y_t} = y_t}$ where $\I$ is an indicator which evaluates to $1$ if the condition is true, and $0$ otherwise. Since the indicator function is not convex, the optimization can be difficult to solve. As a result, a convex loss function is often defined, and is denoted by $\ell(\f(\x), y)$ and the entire optimization can be formulated as:
\begin{eqnarray}
\min_\w \sum_{t=1}^T \ell(\f(\x_t), y_t)
\end{eqnarray}

Several types of loss functions can be used, including the popular hinge-loss $\ell(\f(\x), y) = \frac{1}{2} \max(1 - y\f(\x), 0)$, or the squared-loss $\ell(\f(\x), y) = \frac{1}{2} (\f(\x) - y)^2$.
Sometimes, a regularization term is added to prevent over-fitting or to learn sparse models, or the loss function is modified based on cost-sensitivity of the data (e.g., imbalanced class distribution, different costs for diverse threats).

We will now discuss the existing studies for this task on feature representation and machine learning algorithms design in detail.

\begin{figure*}[ht!]
	\centering
	\includegraphics[width=80mm]{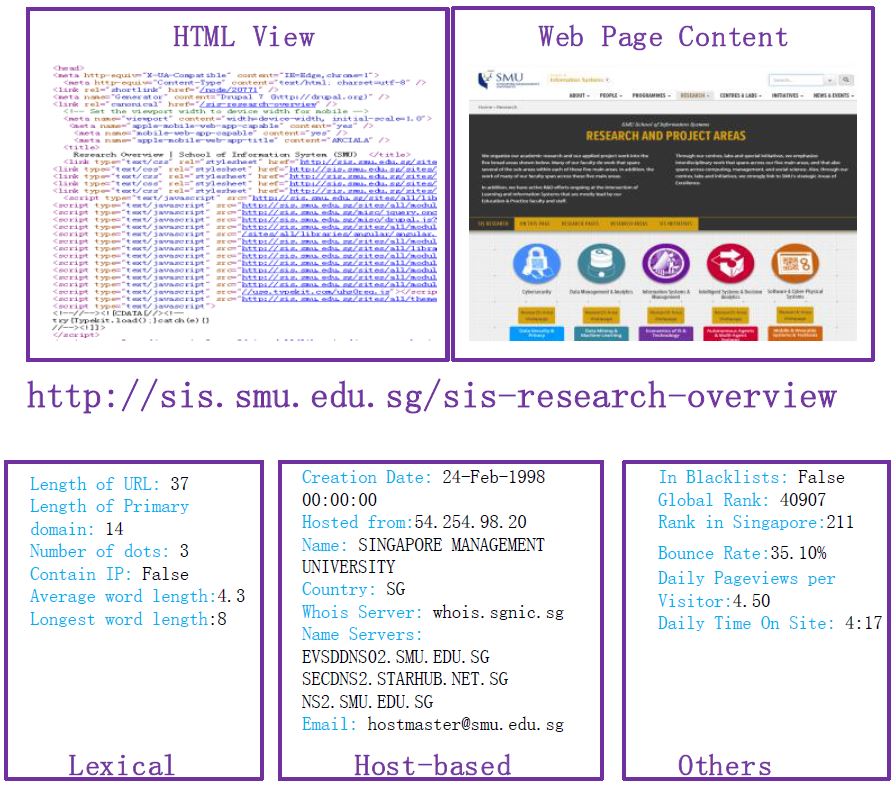}
	\vspace{-0.1in}
	\caption{Example of information about a URL that can be obtained in the Feature Collection stage}
	\label{fig:feature}
	\vspace{-0.2in}
\end{figure*}

\section{Feature Representation}

As stated earlier, the success of a machine learning model critically depends on the quality of the training data, which hinges on the quality of feature representation. Given a URL $\u \in \mathbb{U}$, where $\mathbb{U}$ denotes a domain of any valid URL strings, the goal of feature representation is to find a mapping $g:\mathbb{U}\rightarrow\R^d$, such that $g(\u) \rightarrow \x$ where $\x \in \R^d$ is a $d$-dimensional feature vector, that can be fed into machine learning models. The process of feature representation can be further broken down into two steps:
\begin{enumerate}
	\item \emph{Feature Collection}: This phase is engineering oriented, which aims to collect relevant information about the URL. This includes information such as presence of the URLs in a blacklist, features obtained from the URL String, information about the host, the content of the website such as HTML and JavaScript, popularity information, etc. Figure \ref{fig:feature} gives an example to demonstrate various types various types of information that can be collected from a URL to obtain the feature representation.
	\item \emph{Feature Preprocessing}: In this phase, the unstructured information about the URL (e.g. textual description) is appropriately formatted, and converted to a numerical vector so that it can be fed into machine learning algorithms. For example, the numerical information can be used as is, and Bag-of-words is often used for representing textual or lexical content.
\end{enumerate}

For malicious URL detection, researchers have proposed several types of features that can be used to provide useful information. We categorize these features into: Blacklist Features, URL-based Lexical Features, Host-based features, Content-based Features, and Others (Context and Popularity). All have their benefits and short-comings - while some are very informative, obtaining these features can be very expensive. Similarly, different features have different preprocessing challenges and security concerns. Next, we will discuss each of these feature categories in detail.

\subsection{Blacklist Features}
As mentioned before, a trivial technique to identify malicious URLs is to use blacklists. An existing URL as having been identified as malicious (either through extensive analysis or crowd sourcing) makes its way into the list. However, it has been noted that blacklisting, despite its simplicity and ease of implementation, suffers from nontrivial high false negatives \cite{sinha2008shades} due to the difficulty in maintaining exhaustive up-to-date lists.
Consequently, instead of using blacklist presence alone as a decision maker, it can be used as a powerful feature. In particular, \cite{ma2009beyond} used the presence in a blacklist as a feature, from 6 different blacklist service providers. They also analyzed the effectiveness of these features compared to other features, and observed that blacklist features alone did not have as good a performance as other features, but when used in conjunction with other features, the overall performance of the prediction model improved.

\cite{prakash2010phishnet} observed that to evade detection via blacklisting, many attackers made minor modifications to the original URL. They proposed to extend the blacklist by deriving new URLs based on five heuristics including: Replacing Top-Level Domain (TLDs), IP Address Equivalence, Directory Structure Similarity, Query String substitution, and brand name equivalence. Since even a minor mismatch from the blacklist database can cause a malicious URL to go undetected, they also devised an approximate matching solution. Similar heuristics potentially could be used when deriving blacklist features for machine learning approaches. A similar methodology was adopted for automated URL blacklist generation by \cite{sun2015autoblg,bo2016automating}. \cite{felegyhazi2010potential} developed a method to proactively perform domain blacklisting.



\subsection{Lexical Features}
Lexical features are features obtained from the properties of the URL name (or the URL string). The motivation is that based on how the URL "looks" it should be possible to identify malicious nature of a URL. For example, many obfuscation methods try to "look" like benign URLs by mimicking their names and adding a minor variation to it. In practice, these lexical features are used in conjunction with several other features (e.g. host-based features) to improve model performance. However, using the original URL name directly is not feasible from a machine learning perspective. Instead, the URL string has to be processed to extract useful features. Next, we review the lexical features used for this task.

\emph{Traditional Lexical Features}: The most commonly used lexical features include statistical properties of the URL string, like length of URL, length of each component of the URL (Hostname, Top Level Domain, Primary domain, etc.), number of special characters, etc. \cite{kolari2006svms} were one of the first to suggest extracting words from the URL string. The string was processed such that each segment delimited by a special character (e.g. "/", ".", "?", "=", etc.) comprised a word. Based on all the different types of words in all the URLs, a dictionary was constructed, i.e., each word became a feature. If the word was present in the URL, the value of the feature would be $1$, and $0$ otherwise. This is also known as the bag-of-words model.

Directly using the bag-of-words model, causes a loss of information about the order in which the words occurred in the URL. \cite{ma2009beyond,ma2009identifying} used similar lexical features with modifications to account for the order in which the words appeared. \cite{ma2009identifying} made the distinction between tokens belonging to the hostname, path, top-level domain and primary domain name. This was done by having a separate dictionary for each component. The distinction allowed preserving some of the order in which the words occurred. For example, it enabled distinguishing between the presence of "com" in the top-level domain vs other parts of the URL. \cite{blum2010lexical} enhanced the lexical features by considering the usage of bi-gram features, i.e., they construct a dictionary, where in addition to single-words the presence of a set of 2-words in the same URL is considered a feature. In addition, they recorded the position of sensitive tokens and bigrams to exploit the token context sensitivity.

The entire bag-of-word features approach can be viewed as a machine learning compatible fuzzy blacklist approach. Instead of focusing on the entire URL string, it assigns scores to the URL based on smaller components of the URL string. While this approach offers us an extensive number of features, it can become problematic while running sophisticated algorithms on them. For example, \cite{ma2009identifying} collected a dataset of 2 million URLs, having almost as many lexical features. This number may grow even larger if bi-gram features were considered (where a set of two words became a feature). \cite{kolari2006svms} considered $n$-gram features (same as bi-gram, but $n$ can be $> 2$), and devised a feature selection scheme based on relative entropy to reduce the dimensionality. A similar feature extraction method was used by \cite{zhang2011malicious}, where the feature weights were computed based on the ratio of their presence in one class of URLs against their presence in both classes of URLs.

In order to avoid being caught by blacklists, hackers can generate malicious URLs algorithmically. Using bag-of-words feature for such URLs is likely to give a poor performance, as algorithmically generated URLs may produce never before seen words (hence never before seen features). To detect such algorithmically generated malicious URLs, \cite{yadav2010detecting} analyzed character level strings to obtain the features. They argued that algorithmically generated domain names and those generated by humans would have a substantially different alpha-numeric distribution. Further, since the number of characters is small, the number of features obtained would also be small. They performed their analysis based on KL-divergence, Jaccard Coefficient, and Edit-distance using unigram and bigram distributions of characters.

\emph{Advanced Lexical Features}:
Traditional lexical features were directly obtained from the URL string without significant domain knowledge or computation. Researchers have proposed several advanced lexical features to get more informative features. \cite{le2011phishdef} derive new lexical features using heuristics with the objective of being obfuscation resistant. Based on the obfuscation types identified by \cite{garera2007framework}, five categories of features are proposed: URL-related features (keywords, length, etc.), Domain features (length of domain name, whether IP address is used as domain name, etc.), Directory related features (length of directory, number of subdirectory tokens, etc.), File name features (length of filename, number of delimiters, etc.), and Argument Features(length of the argument, number of variables, etc.).

Another feature is based on the \emph{Kolmogorov Complexity} \cite{pao2012malicious}. Kolmogorov Complexity is a measure of complexity of a string $s$. Conditional Kolmogorov Complexity is the measure of the complexity of a string $s$ given another string for free. This means that the presence of the free string does not add to the complexity of the original input string $s$. Based on this, for a given URL, we compute the URL's Conditional Kolmogorov Complexity with respect to the set of Benign URLs and the set of Malicious URLs. Combining these measures we get a sense of whether the given URL is more similar to the Malicious URL database or the Benign URL database. This feature, though useful, may not be easy to scale up to very large number of URLs. \cite{marchal2014phishscore,marchal2014phishstorm} define a new concept of intra-URL relatedness which is a measure of the relation between different words that comprise the URL with specific focus on relationship between the registered domain and the rest of the URL. \cite{chu2013protect} propose new distance metrics: \emph{domain brand name distance} and \emph{path brand name distance}. These are types of edit distance between strings aimed at detecting those malicious URLs which try to mimic popular brands or websites.

\subsection{Host-based Features}

Host-based features are obtained from the host-name properties of the URL \cite{ma2009identifying}. They allow us to know the location, identity, and the management style and properties of malicious hosts.

\cite{mcgrath2008behind} studied the impact of a few host-based features on the maliciousness of URLs. Some of the key observations were that phishers exploited Short URL services; the time-to-live from registration of the domain was almost immediate for the malicious URLs; and many used botnets to host themselves on multiple machines across several countries. Consequently, host-based features became an important element in detecting malicious URLs.

\cite{ma2009beyond,ma2009identifying} borrowed ideas from \cite{mcgrath2008behind} and proposed the usage of several host-based features including: \emph{IP Address properties}, \emph{WHOIS information}, \emph{Location}, \emph{Domain Name Properties}, and \emph{Connection Speed}. The IP Address properties comprise features obtained from IP address prefix and autonomous system (AS) number. This included whether the IPs of A, MX or NS records are in the same ASes or prefixes as one another. The WHOIS information comprises domain name registration dates, registrars and registrants. The Location information comprises the physical Geographic Location - e.g. country/city to which the IP address belongs. The Domain Name properties comprise time-to-live values, presence of certain keywords like "client" and "server", if the IP address is in the host name or not and does the PTR record resolve one of the host's IP addresses. Since many of the features are identity related information, a bag-of-words like approach is required to store them in a numerical vector, where each word corresponds to a specific identity. Like the lexical features, adopting such an approach leads to a large number of features. For the 2 million URLs, \cite{ma2009identifying} obtained over a million host-based features. Exclusive usage of IP Address Features has also been considered \cite{chiba2012detecting}. IP Address Features are arguably more stable, as it is difficult to obtain new IP Addresses for malicious URLs continuously. Due to this stability, they serve as important features. However, it is cumbersome to use IP Address directly. Instead, it is proposed to extract IP Address features based on a binarization or categorization approach through which octet-based, extended-octet based and bit-string based features are generated.

DNS Fluxiness features were proposed to look for malicious URLs that would hide their identity by using proxy networks and quickly changing their host \cite{holz2008detection,choi2011detecting}. 
\cite{antonakakis2010building} propose using additional host-based features including BGP prefixes and honeypot features. This work is extended in \cite{antonakakis2011detecting} where traffic features are incorporated. \cite{antonakakis2012throw} look into analyzing streams of unsuccessful domain name resolutions.     
\cite{chu2013protect} define \emph{domain age} and \emph{domain confidence} (dependent on similarity with a white-list) level which help determine the fluxiness nature of the URL (e.g. malicious URLs using fast flux will have a small domain age). \cite{sorio2013detection} propose new features to detect malicious URLs that are hidden within trusted sites. They extract \emph{header} features from HTTP response headers. They also use \emph{age} obtained from the time stamp value of the last modified header. \cite{xu2013cross} propose \emph{Application Layer} and \emph{Network Layer} features to devise a cross-layer mechanism to detect malicious URLs. \cite{chiba2016domainprofiler} suggest the usage of \emph{temporal variation patterns} based on active analysis of DNS logs, to help discover domain names that could be abused in the future.

\subsection{Content-based Features}

Content-based features are those obtained upon downloading the entire webpage. As compared to URL-based features, these are "heavy-weight", as a lot of information needs to be extracted, and at the same time, safety concerns may arise. However, with more information available about a particular webpage, it is natural to assume that it would lead to a better prediction model. Further, if the URL-based features fail to detect a malicious URL, a more thorough analysis of the content-based features may help in early detection of threats \cite{canali2011prophiler}. The content-based features of a webpage can be drawn primarily from its HTML content, and the usage of JavaScript. \cite{hou2010malicious} categorize the content based features of a webpage into 5 broad segments: Lexical features, HTML Document Level Features, JavaScript features, ActiveX Objects and feature relationships. \cite{zhang2007cantina,xiang2011cantina+} proposed CANTINA and its variants for detecting phishing websites using a comprehensive feature-based machine learning approach, by exploiting various features from the HTML Document Object Model (DOM), search engines and third party services. In the following we discuss some of these categories, primarily focusing on the HTML Document Level Features and JavaScript Features. We also discuss Visual Features where the properties of the images of the webpages are used to detect the maliciousness of a URL.

\subsubsection{HTML Features}

\cite{hou2010malicious} proposed the usage of lexical features from the HTML of the webpage. These are relatively easy to extract and preprocess. At the next level of complexity, the HTML document level features can be used. The document level features correspond to the statistical properties of the HTML document, and the usage of specific types of functionality. \cite{hou2010malicious} propose the usage of features like: length of the document, average length of the words, word count, distinct word count, word count in a line, the number of NULL characters, usage of string concatenation, unsymmetrical HTML tags, the link to remote source of scripts, and invisible objects. Often malicious code is encrypted in the HTML, which is linked to a large word length, or heavy usage of string concatenation, and thus these features can help in detecting malicious activity.

Similar features with minor variations were used by many of the subsequent researchers including \cite{choi2011detecting} (number of iframes, number of zero size iframes, number of lines, number of hyperlinks, etc.). \cite{canali2011prophiler} also used similar features, and additionally proposed to use several more descriptive features which were aimed at minor statistical properties of the page. These include features such as number of elements with a small area, number of elements with suspicious content (suspiciousness was determined by the length of the content between the start and end tag), number of out of place elements, presence of double documents, etc. \cite{borgolte2013delta} developed a \emph{delta} method, where \emph{delta} represented the change in different versions of the website. They analyzed whether the change was malicious or benign.

\subsubsection{JavaScript Features}

\cite{hou2010malicious} argue that several JavaScript functions are commonly used by hackers to encrypt malicious code, or to execute unwanted routines without the client's permission. For example extensive usage of function \emph{eval()} and \emph{unescape()} may indicate execution of encrypted code within the HTML. They aim to use the count of 154 native JavaScript functions as features to identify malicious URLs. \cite{choi2011detecting} identify a subset (seven) of these native JavaScript functions that are often in Cross-site scripting and Web-based malware distribution. These include: escape(), eval(), link(), unescape(), exec(), and search() functions. \cite{canali2011prophiler} propose additional heuristic JavaScript features including: keywords-to-words ratio, number of long strings, presence of decoding routines, shell code presence probability, number of direct string assignments, number of DOM-modifying functions, number of event attachments, number of suspicious object names, number of suspicious strings, number of "iframe" strings and number of suspicious string tags.
In \cite{choi2010automatic}, the authors try to detect JavaScript Obfuscation by analyzing the JavaScript codes using n-gram, Entropy and Word Size. n-gram and word size are commonly used to look for character/word distribution and presence for long strings. For Entropy of the strings, they observe that obfuscated strings tend to have a lower entropy. More recently, \cite{wang2016deep} applied deep learning techniques to learn feature representations from JavaScript code.

\subsubsection{Visual Features}

There have also been attempts made at using images of the webpages to identify the malicious nature of the URL. Most of these focus on computing visual similarity with \emph{protected pages}, where the protected pages refer to genuine websites. Finding a high level of visual similarity of a suspected malicious URL could be indicative of an attempt at phishing. One of the earliest attempts at using visual features for this task was by computing the Earth Mover's Distance between 2 images \cite{fu2006detecting}. \cite{wenyin2005detection,liu2006antiphishing} addressed the same problem and developed a system to extract visual features of webpages based on text-block features and image-block features (using information such as block size, color, etc.). More advanced computer vision technologies were adapted for this task. Contrast Context Histogram (CCH) features were suggested \cite{chen2009fighting}, and so were Scale Invariant Feature Transform (SIFT) features \cite{afroz2011phishzoo}. Another approach using visual feature was developed by \cite{dunlop2010goldphish}, where an OCR was used to read the text in the image of the webpage.
\cite{zhang2011textual} combine both textual and visual features for measuring similarity.
With recent advances in Deep Learning for Image Recognition \cite{krizhevsky2012imagenet,he2015deep}, it may be possible to extract more powerful and effective visual features.

\subsubsection{Other Content-based Features}
\cite{hou2010malicious} argued that due to the powerful functionality of ActiveX objects, they can be used to create malicious DHTML pages. Thus, they tried to compute frequency for each of eight ActiveX objects. Examples include: ``Scripting.FileSystemObject" which can be used for file system I/O operations, ``WScript.Shell" which can execute shell scripts on the client's computer, and ``Adodb.Stream" which can download files from the Internet.
\cite{xiang2009hybrid} try to find the identity and keywords in the DOM text and evaluate the consistency between the identity observed and the identity it is potentially trying to mimic which is found by searching.
\cite{soska2014automatically} used the directory structure of the websites to obtain insights.

\subsection{Other Features}

\subsubsection{Context Features}
Recent years have seen the growth of Short URL service providers, which allow the original URL to be represented by a shorter string. This enables sharing of the URLs in on social media platforms like twitter, where the originally long URLs would not fit within the 140 character limit of a tweet. Unfortunately, this has also become a popular obfuscation technique for malicious URLs. While the Short URL service providers try their best to not generate short URLs for the malicious ones, they struggle to do an effective job \cite{maggi2013two,gupta2014bit}. As a result, a recently emerging research direction has become active where \emph{context-features} of the URL are obtained, i.e., the features of the background information where the URL has been shared. \cite{lee2012warningbird} use context information derived from the tweets where the URL was shared. \cite{wang2013click} used click traffic data to classify short URLs as malicious or not. \cite{cao2016detection} propose forwarding based features to combat forwarding-based malicious URLs. \cite{cao2015detecting} propose another direction of features to identify malicious URLs - they also focus on URLs shared on social media, and aim to identify the malicious nature of a URL by performing behavioral analysis of the users who shared them, and the users who clicked on them. These features are formally called "Posting-based" features and "Click-based" features. \cite{alshboul2015detecting} approach this problem with a systematic categorization of context features which include content-related features (lexical and statistical properties of the tweet), context of the tweet features (time, relevance, and user mentions) and social features (following, followers, location, tweets, retweets and favorite count).

\subsubsection{Popularity Features}
Some other features used were designed as heuristics to measure the \emph{popularity} of the URL. One of the earliest approaches to applying statistical techniques to detect malicious URLs \cite{garera2007framework} aimed at probabilistically identifying the importance of specific hand-designed features. These include Page-based features (Page rank, quality, etc.), Domain-based features (presence in \emph{white domain table}), Type-based features (obfuscation types) and Word-based features(presence of keywords such as "confirm", "banking", etc.). \cite{thomas2011design} use both the URL-based and content based features, and additionally record the initial URL, the landing URL and the redirect chain. Further they record the number of popups and the behavior of plugins, which have been commonly used by spammers. \cite{choi2011detecting} proposed the usage of new categories of features: Link Popularity and Network Features. Link Popularity is scored on the basis of incoming links from other webpages. This information was obtained from different search engines. In order to make the usage of these features robust to manipulation, they also propose the usage of certain metrics that validate the quality of the links. They also use a metric to detect spam-to-spam URL links. For their work, they use these features in conjunction with lexical features content-based feature, and host-based features. \cite{eshete2013binspect} used social reputation features of URLs by tracking their public share count on Facebook and Twitter. Other efforts  incorporated information on redirection chains into redirection graphs, which provided insight into detecting malicious URLs \cite{stringhini2013shady,kwon2017domain}.
\cite{marchal2014phishstorm,ding2019keyword} use search engine query data to mine for word relatedness measurement.

\subsection{Summary of Feature Representations}
There is a wide variety of information that can be obtained for a URL. Crawling the information and transforming the unstructured information to a machine learning compatible feature vector can be very resource intensive. While extra information can improve predictive models (subject to appropriate regularization), it is often not practical to obtain a lot of features. For example, several host-based features may take a few seconds to be obtained, and that itself makes using them in real world setting impractical. Another example is the Kolmogorov Complexity - which requires comparing a URL to several malicious and benign URLs in a database, which is infeasible for comparing with billions of URLs. Accordingly, care must be taken while designing a Malicious URL Detection System to tradeoff the usefulness of a feature and the difficulty in retrieving it. We present a subjective evaluation of different features used in literature. Specifically, we evaluate them on the basis of Collection Difficulty, Associated Security Risks, need for an external dependency to acquire information, the associated time cost with regard to feature collection and feature preprocessing, and the dimensionality of the features obtained.

Collection difficulty is refers to the engineering effort required to obtain specific information about the features. Blacklist, context and popularity features require additional dependencies and thus have a higher collection overhead, whereas the other features are directly obtained from the URL itself. This also implies, that for a live-system (i.e. real-time Malicious URL Detection), obtaining features with a high collection time may be infeasible. In terms of associated security risks, the content-features have the highest risk, as potential malware may be explicitly downloaded while trying to obtain these features, while other features do not suffer from these issues. The collection time of the blacklist features can be high if the external dependency has to be queried during runtime, however, if the the entire blacklist can be stored locally, the collection overhead is very small. Collection of the lexical features is very efficient, as they are basically direct derivatives of the URL string. Host-based features are relatively time-consuming to obtain. Content-features usually require downloading the webpage which would affect the feature collection time. For preprocessing, once the data has been collected, deriving the features in most cases is computationally very fast. For dimensionality size, the lexical features have a very high-dimensionality (and so do unstructured Host-features and content features). This is largely because they are all stored as Bag-of-Words features. This feature size consequently affects the training and test-time. These properties are summarized in Table \ref{tab:featureComparison}. We also categorize the representative references according to the feature representation used, in Table \ref{tab:features}.

\begin{table*}[htbp]
	\small
	\caption{Properties of different Feature Representations for Malicious URL Detection}
	\centering
	\begin{tabular}{|c|c|cccccc|}
		\hline
		\multicolumn{1}{|l|}{\textbf{Features}} & \textbf{Category} &  \multicolumn{6}{c|}{\textbf{Criteria}} \\
		\cline{3-8}
		&       &  \textbf{Collection}     &   \textbf{}    &  \textbf{External}     & \textbf{Collection} & \textbf{Processing} & \textbf{Feature}  \\
		&        & \textbf{Difficulty} & \textbf{Risk} & \textbf{Dependency} & \textbf{Time } & \textbf{ Time} & \textbf{Size} \\
		\hline
		\hline
		\multicolumn{1}{|l|}{\textbf{Blacklist}} & \textbf{Blacklist}  & Moderate & Low   & Yes   & Moderate & Low   & Low \\
		\hline
		\multirow{2}[0]{*}{\textbf{Lexical}} & \textbf{Traditional}  & Easy  & Low   & No    & Low   & Low   & Very High \\
		& \textbf{Advanced} &  Easy  & Low   & No    & Low   & High  & Low \\
		\hline
		\multirow{2}[0]{*}{\textbf{Host}} & \textbf{Unstructured} & Easy  & Low   & No    & High  & Low   & Very High \\
		& \textbf{Structured}  & Easy  & Low   & No    & High  & Low   & Low \\
		\hline
		\multirow{4}[0]{*}{\textbf{Content}}& \textbf{HTML}& Easy  & High  & No    & Depends & Low   & High \\
		& \textbf{JavaScript} & Easy  & High  & No    & Depends & Low   & Moderate \\
		& \textbf{Visual} & Easy  & High  & No    & Depends & High   & High \\
		& \textbf{Other}  & Easy  & High  & No    & Depends & Low   & Low \\
		\hline
		\multirow{2}[0]{*}{\textbf{Others}} & \textbf{Context}  & Difficult & Low   & Yes   & High  & Low   & Low \\
		& \textbf{Popularity}  & Difficult & Low   & Yes   & High  & Low   & Low \\
		\hline
	\end{tabular}%
	\label{tab:featureComparison}%
\end{table*}%

\begin{table*}[htbp]
	\small
	\caption{Representative references of different types of features used by researchers in literature}
	\centering
	\begin{tabular}{|c|c|c|}
		\hline
		\textbf{Feature} & \textbf{Sub Category} & \textbf{Representative References}\\
		\hline
		\hline
		\textbf{Blacklist} & Blacklist &    \cite{garera2007framework,ma2009beyond,prakash2010phishnet,felegyhazi2010potential,sun2015autoblg,bo2016automating}      \\
		\hline
		\multirow{3}[0]{*}{{\textbf{Lexical}}}  & \multirow{3}[0]{*}{{Lexical}}  &    \cite{kolari2006svms,garera2007framework,ma2009beyond,ma2009identifying,he2010mining,blum2010lexical,ma2010exploiting,whittaker2010large,yadav2010detecting,gastellier2011decisive,zhang2011textual,bannur2011judging,bilge2011exposure,canali2011prophiler,choi2011detecting,ma2011learning,zhang2011malicious}\\
		&& \cite{le2011phishdef,thomas2011design,huang2012svm,pao2012malicious,maurer2012sophisticated,chu2013protect,eshete2013binspect,eshete2013einspect,eshete2013effective,lin2013malicious,sorio2013detection,ranganayakulu2013detecting,su2013suspicious,xu2013cross,huang2014malicious}\\
		&&
		\cite{marchal2014phishscore,wang2015breaking,marchal2014phishstorm,marchal2015know,sato2016,jain2017towards,verma2017s,saxe2017expose,rajitha2017suspicious,buber2017nlp,shima2018classification,le2018urlnet,patil2018malicious,yu2018character, sahingoz2019machine,li2019stacking,tupsamudre2019everything}\\
		\hline
		\textbf{Host} & Host-based &   \cite{mcgrath2008behind,ma2009beyond,ma2009identifying,ma2010exploiting,whittaker2010large,bilge2011exposure,canali2011prophiler,choi2011detecting,ma2011learning,thomas2011design,antonakakis2010building,antonakakis2011detecting,antonakakis2012throw,holz2008detection,ranganayakulu2013detecting,xu2013cross,gugelmann2015automated,kuyama2016method,rajitha2017suspicious,tan2018domainobserver}\\
		\hline
		\multirow{5}[0]{*}{\textbf{Content}} & \multirow{2}[0]{*}{{HTML}}  &   \cite{pan2006anomaly,seifert2008identification,liang2009malicious,hou2010malicious,bannur2011judging,he2011efficient,gastellier2011decisive,choi2011detecting,canali2011prophiler,thomas2011design}\\
		&& \cite{eshete2013binspect,eshete2013einspect,eshete2013effective,xu2013cross,yoo2014two,marchal2015know,zhang2016two,canfora2016set,jain2017towards,altay2018context,jain2019machine,mao2019phishing,li2019stacking}\\
		\cline{2-3}
		\textbf{} & JavaScript  &   \cite{liang2009malicious,hou2010malicious,choi2010automatic,cova2010detection,canali2011prophiler,kim2011suspicious,choi2011detecting,thomas2011design,eshete2013binspect,eshete2013einspect,eshete2013effective,xu2013cross,yoo2014two,wang2016deep}\\
		\cline{2-3}
		\textbf{} & Visual &   \cite{fu2006detecting,wenyin2005detection,liu2006antiphishing,chen2009fighting,zhang2011textual,bannur2011judging,afroz2011phishzoo,dunlop2010goldphish,hara2009visual}    \\
		\cline{2-3}
		\textbf{} & Others &   \cite{xiang2009hybrid,hou2010malicious,yoo2014two,soska2014automatically}    \\
		\hline
		\multirow{2}[0]{*}{\textbf{Others}} & Context-based  &  \cite{lee2012warningbird,aggarwal2012phishari,wang2013click,cao2016detection,cao2015detecting,alshboul2015detecting,nepali2016you}\\
		\cline{2-3}
		& Popularity-based  &  \cite{garera2007framework,whittaker2010large,wenyin2010discovering,choi2011detecting,huh2011phishing,sunil2012pagerank,eshete2013binspect,eshete2013einspect,eshete2013effective,stringhini2013shady,marchal2014phishstorm,hu2016identifying,ding2019keyword}\\
		\hline
	\end{tabular}%
	\label{tab:features}
\end{table*}%


\section{Machine Learning Algorithms for Malicious URL Detection}

\label{sec:ml}
There is a rich family of machine learning algorithms in literature, which can be applied for solving malicious URL detection. After converting URLs into feature vectors, many of these learning algorithms can be generally applied to train a predictive model in a fairly straightforward manner. However, to effectively solve the problem, some efforts have also been explored in devising specific learning algorithms that either exploit the properties exhibited by the training data of Malicious URLs, or address some specific challenges which the application faces. In this section, we categorize and review the learning algorithms that have been applied for this task, and also suggest suitable machine learning technologies that can be used to solve specific challenges encountered.
We categorize the learning algorithms into: Batch Learning Algorithms, Online Algorithms, Representation Learning, and Others. Batch Learning algorithms work under the assumption that the entire training data is available prior to the training task. Online Learning algorithms treat the data as a stream of instances, and learn a prediction model by sequentially making predictions and updates. This makes them extremely scalable compared to batch algorithms. We also discuss the extensions of Online Learning to cost-sensitive and active learning scenarios. Next, we discuss representation learning methods, which are further categorized into Deep Learning and Feature Selection techniques. Lastly, we discuss other learning algorithms, in challenges specific to Malicious URL Detection are addressed, including unsupervised learning, similarity learning, and string pattern matching.

\subsection{Batch Learning}
Following the previous problem setting, consider a URL data set with $T$ URLs $\{(\u_1, y_1), \dots, (\u_T, y_T)\}$, where $\u_t\in\mathbb{U}$ for $t\in 1,\dots, T$ represents a URL from the training data, and $y_t \in \{1, -1\}$ is its class label where $y = 1$ indicates a malicious URL and $y=-1$ indicates a benign URL. Using an appropriate feature representation scheme ($g:\mathbb{U}\mapsto\mathbb{R}^d$) as discussed in the previous section, one can map a URL instance into a $d$-dimensional feature vector, i.e., $g(\u_t)\rightarrow\x_t$. As a result, one can apply any existing learning algorithm that can work with vector space data to train a predictive model for malicious URL detection tasks. In this section we review the most common batch learning algorithms that have been applied for Malicious URL Detection.

A family of batch learning algorithms falls under a discriminative learning framework using regularized loss minimization as:
\begin{eqnarray}
\min_\f \sum_{t=1}^T \ell(\f(\x_t), y_t) + \lambda\mathcal{R}(\mathbf{w})
\end{eqnarray}
where $\f(\x_t)$ can be either a linear model, e.g., $\f(\x_t)=\w\cdot\x_t + b$, or some nonlinear models (kernel-based or neural networks), $\ell(\f(\x_t), y_t)$ is some loss function to measure the difference between the model's prediction $\f(\x_t)$ and the true class label $y$, $\mathcal{R}(\mathbf{w})$ is a regularization term to prevent overfitting, and $\lambda$ is a regularization parameter to trade-off model complexity and simplicity. In the following, we discuss two popular learning algorithms under this framework: Support Vector Machines and Logistic Regression.

\subsubsection{Support Vector Machine}
(SVM) is one of most popular supervised learning methods. It exploits the structural risk minimization principle using a maximum margin learning approach, which essentially can be viewed as a special instance of the regularized loss minimization framework. Specifically, by choosing the hinge loss as the loss function and maximizing the margin, SVM can be formulated into the following optimization:
\begin{eqnarray}
\nonumber
(\w, b) \leftarrow \underset{\w, b}{\text{arg min}} \frac{1}{T} \sum_{t=1}^T  \max(0, 1 - y_t (\w \cdot \x_t + b)) + \lambda \|\w\|_2^2
\end{eqnarray}
In addition, SVM can learn nonlinear classifiers using kernels \cite{Smola1998}.
SVMs are probably one of the most commonly used classifiers for Malicious URL Detection in literature  \cite{kolari2006svms,pan2006anomaly,ma2009beyond,hou2010malicious,bannur2011judging,he2011efficient,huang2012svm,pao2012malicious,chu2013protect,sorio2013detection,wang2013click,xu2013cross,marchal2014phishscore,marchal2014phishstorm,alshboul2015detecting}.

\subsubsection{Logistic Regression}
is another well-known discriminative model which computes the conditional probability for a feature vector $\x$ to be classified as a class $y=1$ by
\begin{eqnarray}
P(y=1 | \x; \w, b) = \sigma(\w \cdot \x + b) = \frac{1}{1+e^{-(\mathbf{w}\cdot\mathbf{x}+b)}}
\end{eqnarray}
Based on the maximum-likelihood estimation (equivalently defining the loss function as the negative log likelihood), the optimization of logistic regression can be formulated as
\begin{eqnarray}
(\w, b) \leftarrow \underset{\w, b}{\text{arg min}}\frac{1}{T} \sum_{t=1}^T -\log P(y_t | \x_t; \w,b)  + \lambda \mathcal{R}(\w)
\end{eqnarray}
where the regularization term can be either L2-norm $\mathcal{R}(\w) = ||\w||_2$ or L1-norm $\mathcal{R}(\w) = ||\w||_1$ for achieving a sparse model for high-dimensional data.
Logistic Regression has been a popular learning method for Malicious URL Detection \cite{garera2007framework,ma2009beyond,bannur2011judging,canali2011prophiler,lee2012warningbird,wang2013click,xu2013cross}.

Other commonly used supervised learning algorithms focus on feature-wise analysis to obtain the prediction model. These include the Naive Bayes Classifier which computes the posterior probability of the class label assuming feature independence, and Decision Tree which adopts a greedy approach to constructing if-else rules based on the features offering the best splitting criteria.

\subsubsection{Naive Bayes}
is a generative model for classification, which is ``naive" in the sense that this model assumes all features of $\x$ are independent of each other. Specifically, let $P(\x|y)$ denote the conditional probability of the feature vector given a class, the independence assumption implies that $P(\x | y) = \Pi_{i=1}^d P(x_i | y)$, where $d$ is the number of features. By applying the Bayes Theorem, one can compute the posterior probability that a feature vector $\x$ is a malicious URL by
\begin{eqnarray}
P(y=1 | \x) = \frac{P(\x|y = 1)}{P(\x | y = 1) + P(\x | y = -1)}
\end{eqnarray}
Naive Bayes has been used for Malicious URL Detection by \cite{hou2010malicious,canali2011prophiler,zhang2011textual,aggarwal2012phishari,xu2013cross,cao2016detection}.

\subsubsection{Decision Trees}
is one of most popular methods for inductive inference and has a major advantage of its highly interpretable decision tree classification models which can also been converted into a rule set for human readability.
Decision Trees have been used for malicious URL/web classification by \cite{seifert2008identification,bilge2011exposure,canali2011prophiler,aggarwal2012phishari,chiba2012detecting,wang2013click,xu2013cross,cao2016detection,marchal2014phishscore,marchal2014phishstorm,alshboul2015detecting}.
A closely related approach which gives us rules in the form of If-then was applied in using Associative Classification mining by \cite{abdelhamid2014phishing}.

\subsubsection{Others and Ensembles}

In addition to the above, other recently proposed approaches include applying Extreme Learning Machines (ELM) for classifying the phishing web sites  \cite{zhang2016two}. The spherical classification approach that allows batch learning models to be suitable for a large number of instances was also used for this task \cite{astorino2016malicious}. Beyond the binary classification approaches, \cite{choi2011detecting} formulated the problem of malicious URL detection as a \emph{multi-label classification} task. The argument for the need of multi-label classification is that different attacks have varying degrees of threats. For example, a spam URL is not as deadly as a malware infection. They proposed a two-step method: first using SVM for classifying a URL as malicious or benign; and second, performing multi-label classification on the malicious URLs using some popular multi-label learning methods (e.g., RAkEL and ML-kNN).

In addition, there are quite a few malicious URL detection approaches using ensemble learning methods. For example, \cite{ramanathan2012phishing} applied Adaboost for detecting phishing websites using a content-based approaching together with Latent Dirichlet Allocation (LDA) for topic modeling. \cite{eshete2013binspect} employed an ensemble of multiple classifiers to make a weighted prediction. They independently train Decision Trees, Random Forests, Bayesian classifiers, Support Vector Machines and Logistic Regression, and design a confidence weighted majority voting scheme to make the final prediction. \cite{su2013suspicious} adopt a multi-view analysis where a logistic regression model is trained on different portions of the URL lexical features, and their optimal combination is learnt. \cite{eshete2013einspect,eshete2013effective} adopt an evolutionary optimization method to search for the best combination of features and models to obtain the final ensemble. In practice, ensemble learning is a common and very successful learning strategy when there is a need for boosting the predictive performance.

Finally, the recent years have witnessed several efforts towards deep learning for malicious URL detection. Despite being batch models, their fundamental goal is to \textit{learn} feature representations, and are thus discussed in Section \ref{sec:representation}. Although batch learning algorithms are popular and easy to use, they can suffer from several major limitations when dealing with real-world malicious URL detection tasks. For example, batch learning methods often suffer from expensive retraining cost when the new training data may arrive frequently. Moreover, due to expensive retraining cost, batch learning algorithms often do not update the model frequently, making them difficult to capture some emerging threats in a timely way. Last but not least, batch learning methods may poorly adapt to the concept drift due to their nature of batch training. To address these limitations, online learning algorithms have been emerging as a promising direction for resolving the Malicious URL Detection tasks.


\subsection{Online Learning}
Online Learning represents a family of efficient and scalable learning algorithms that learn from data sequentially \cite{Cesa-Bianchi2006,hoi2014libol,hoi2018online}. Consider malicious URL detection, given a sequence of $T$ labeled instances, denoted by $\D = \{(\x_1, y_1), \dots, (\x_T, y_T)\}$, where $\x_t \in \R^d$ denotes the URL's feature representation, and $y_t \in \{-1, +1\}$ is the class label. $y=+1$ denotes a malicious URL, and $y_t = -1$ denotes a benign URL. At each iteration $t$, the algorithm makes a prediction $\f(\x_t) = sgn(\w\cdot\x_t)$ where $\w$ is a $d$-dimensional weight vector initialized to $\mathbf{0}$ at $t=0$. After the prediction, the true class label $y_t$ is revealed to the learner, and based on the loss suffered, the learner makes an update of the model in order to improve predictions in the future. The general framework of an online learning algorithm is outlined in Algorithm \ref{alg:online}.
\begin{algorithm}[hptb]
	\caption{The Online Learning Procedure}
	\label{alg:online}
	\begin{algorithmic}
		\STATE Initialize the prediction function as $\w_1 = \mathbf{0}$;
		\FOR{ $t=1,2,\ldots,T$}
		\STATE Receive instance: $\x_t\in \R^d$;
		\STATE Predict $\hat{y_t}= \f_t(\x_t) ( = \sign(\w_t^\top\x_t)$ for binary classification);
		\STATE Receive correct label: $y_t\in\{-1,+1\}$;
		\STATE Suffer loss: $\ell_t(\w_t)$, which depends on the difference between $\w_t^\top\x_t$ and  $y_t$;
		\STATE Update the prediction function $\w_t$ to $\w_{t+1}$;
		\ENDFOR
	\end{algorithmic}
\end{algorithm}
Online learning algorithms are often much more scalable than traditional batch learning algorithms. Both the learning (model updates) and forecasting are computationally very efficient, making it especially suitable for malicious URL detection tasks with increasingly large amounts of training data (often with millions of instances and millions of features), where batch learning algorithms may suffer due to their expensive re-training and the high memory and computational constraints.
Online learning algorithms are often developed with strong theoretical guarantees such that they are able to asymptotically learn the prediction models as good as the batch algorithms under mild assumptions.

Online learning has been actively explored and applied to resolve the malicious URL Detection tasks \cite{ma2009identifying}. In the following, we categorize the existing online learning algorithms roughly into two major categories: (i) First-order online algorithms, and (ii) Second-order online algorithms, and highlight some important concerns for their applications to malicious URL detection.

\subsubsection{First Order Online Learning}
First-order algorithms learn by updating the weight vector $\w$ for classification sequentially by utilizing only the first-order information with training data. We briefly describe some popular first-order online algorithms applied to Malicious URL Detection.

\emph{Perceptron} \cite{Rosenblatt1958} is the earliest online learning algorithm. In each iteration, whenever a mistake is made by the prediction model, Perceptron makes an update as follows:
\begin{eqnarray}
\w_{t+1} \leftarrow \w_t + y_t\x_t
\end{eqnarray}

\emph{Online Gradient Descent} (OGD) \cite{Zinkevich2003} updates the weight vector $\w$ by applying the (Stochastic) Gradient Descent principle only to a single training instance arriving sequentially. Specifically, OGD makes an online update iteratively as: 
\begin{eqnarray}
\w_{t+1} \leftarrow \w_t - \eta \nabla\ell(\w_t, \x_t; y_t)
\end{eqnarray}
where $\eta$ is a step size parameter, and $\ell(\w_t, \x_t; y_t)$ is some predefined loss function, e.g., Hinge-Loss, Negative Log-Likelihood, Squared Loss, etc.

\emph{Passive-Aggressive} learning (PA) \cite{Crammer2006} is an online learning method that trades off two concerns: (i) passiveness: to avoid the new model deviating too much from the existing one, and (ii) aggressiveness: to update the model by correcting the prediction mistake as much as possible.
The optimization of PA learning can be cast as follows:
\begin{eqnarray}
\w_{t+1} \leftarrow \underset{\w}{\text{argmin}} \frac{1}{2}||\w_t - \w||^2  \text{\quad subject to } y_t (\w \cdot \x_t ) \ge 1
\end{eqnarray}
The closed-form solution to the above can be derived as the following update rule:
\begin{eqnarray}
\nonumber
&\w_{t+1} \leftarrow \w_t + \tau_t y_t \x_t \\
\text{\quad where \quad} & \tau_t = \max\big(\frac{1 - y_t (\w_t \cdot \x_t)}{||\x_t||^2},0\big)
\end{eqnarray}
The above model assumes a hard margin exists, that is, data can be linearly separable, which may not be always true, especially when data is noisy. To overcome this limitation, soft-margin PA variants, such as PA-I and PA-II, are often commonly used, which also have closed-form solutions \cite{Crammer2006}.

The above first-order online learning algorithms have been widely applied for malicious URL detection tasks in literature \cite{ma2009identifying,ma2011learning,thomas2011design,zhang2011malicious,feroz2014examination}, which are efficient, scalable, simple to understand, and easy to implement.

\subsubsection{Second Order Online Learning}

Unlike the first order online learning, second order online learning aims to boost the learning efficacy by exploiting second-order information, e.g., the second order statistics of underlying distributions. For example, they usually assume the weight vector $\w$ follows a Gaussian distribution $\w \sim \N(\boldsymbol{\mu}, \Sigma)$ with mean vector $\boldsymbol\mu \in \R^d$ and covariance matrix $\Sigma \in \R^{d \times d}$. This is particularly useful for malicious URL Detection where data is sparse and high dimensional (due to the bag-of-words or alike representations of lexical features). Below we briefly describe some popular second-order algorithms applied to Malicious URL Detection.

\emph{Confidence-Weighted} learning (CW) \cite{dredze2008confidence} is similar to the PA learning algorithms in terms of passiveness and aggressiveness tradeoff, except that CW exploits the second-order information. In particular, CW learning maintains a different confidence measure for each individual feature, such that weights of lower confidence will be updated more aggressively than those of higher confidence.
Specifically, by modeling the weight vector as a Gaussian distribution, CW trades off between (i) passiveness: to avoid the new distribution of the model from deviating too much from the existing one; and (ii) aggressiveness: to update the model by not only correcting the prediction mistake if any, but also improving the classification confidence. More formally, the CW learning can be cast into the following optimization:
\begin{eqnarray}
(\boldsymbol\mu_{t+1}, \Sigma_{t+1}) \leftarrow \underset{\boldsymbol\mu,\Sigma}{\text{argmin }} D_{KL}(\N(\boldsymbol\mu, \Sigma)||\N(\boldsymbol\mu_t, \Sigma_t))\\
\text{subject to } y_t(\boldsymbol\mu, \x_t) \ge \phi^{-1}(\eta)\sqrt{\x_t^\top\Sigma\x_t}
\end{eqnarray}

Like the PA algorithms, the closed-form solutions for the CW optimization can be derived. CW algorithms have been applied for detecting malicious URLs by \cite{ma2009identifying,blum2010lexical}.

CW online learning and its variants have been explored for malicious URL detection in literature. For example, \cite{ma2010exploiting} applied the CW learning for malicious URL detection by improving the efficiency when exploiting the full covariance matrix for high-dimensional features, which uses an approximation technique to accelerate the covariance computation (although it may be still quite slow for very high-dimensional data). Further, Adaptive Regularization of Weights (AROW) \cite{crammer2009adaptive}, an improved CW learning algorithm, for learning with non-separable data, was also used for Malicious URL Detection in \cite{le2011phishdef}. \cite{lin2013malicious} adopted a hybrid online learning technique by combining both CW and PA algorithms. Specifically, CW is used for learning from purely lexical features (e.g., bag of words), and PA is used for learning from descriptive features (e.g., statistical properties of the lexical features). They assume lexical features are more effective at detecting maliciousness, while they could change frequently (short-lived), whereas descriptive properties are more stable and static.

Besides, there are many other kinds of online learning algorithms (both first-order and second-order) in literature \cite{hoi2014libol}, which may also be applicable to Malicious URL Detection, but yet to be extensively studied.


\subsubsection{Cost-Sensitive Online Learning}

Unlike a regular binary classification task, Malicious URL Detection often faces the challenges of imbalanced label distribution (i.e., different number of malicious and benign URLs), and also a differential misclassification cost (e.g. malware installation is much more severe than a simple spam). Accordingly, the designed learning algorithms have to account for this differential misclassification rate in the optimization problem. There have been several algorithms in literature (for both batch and online settings) which address this issue. An example is Cost-Sensitive Online Learning \cite{WZH14,zhao2013cost,sdm-zhao2013cost,zhao2015cost,sahoo2016cost,maurya2016online}. While traditional online learning algorithms simply optimize the classification accuracy or mistake rate (which could be misleading for an extremely imbalanced data set since a trivial algorithm that declares every URL as benign may achieve a very high accuracy), cost-sensitive online learning aims to optimize either one of two cost-sensitive measures: \emph{sum} and \emph{cost}, where \emph{sum} is a weighted combination of specificity and sensitivity, and \emph{cost} is the weighted summation of misclassification costs on positive and negative instances. By defining cost-sensitive loss functions, cost-sensitive online learning algorithms can be derived by applying the similar techniques (e.g., online gradient descent). There have been efforts to address cost-sensitive learning in the batch setting too.

\subsubsection{Online Active Learning}

Traditional supervised learning (either batch or online) methods often assume labels of training data can always be obtained by the learners at no cost. This is not realistic in real systems since labeling data can be quite expensive and time-consuming. Online Active Learning aims to develop an online learning algorithm for training a model that queries the label of an incoming unlabeled URL instance only if there is a need (e.g., according to some uncertainty measure)\cite{lu2016online,icdm2016soal,sculley2007online}. Typically, an active learner works in an online learning manner for a real system. For example, \cite{zhao2013cost,lu2016online} proposed a cost-sensitive online active learning (CSOAL) approach for Malicious URL detection, where the online learner decides to query the label on the fly for an incoming unlabeled URL instance, such that the label will be queried only when the classification confidence of the URL instance is low or equivalently there is a high uncertainty for making a correct prediction.

\subsection{Representation Learning}
\label{sec:representation}

The task of Malicious URL Detection uses a large number of and a variety of features for building prediction models. Based on the context, amount of training data available, and several other factors, determining which features can be useful is a very challenging task. Moreover, it can be difficult (even for domain) experts to identify the best features for the task. If not addressed, such factors can lead to severe overfitting, ineffective models, noisy models and computational burdens (both in terms of speed and required memory). There are two broad categories of work for representation learning for Malicious URL Detection that we review here: i) Deep Learning for Malicious URL Detection; and ii) Feature Selection and Sparsity Regularization Methods. 

\subsubsection{Deep Learning for Malicious URL Detection}
Deep Learning \cite{lecun2015deep} has become a panacea for several data mining and machine learning applications in the recent years, with unprecedented success in (computer vision \cite{krizhevsky2012imagenet}, natural language processing (NLP) \cite{zhang2015character}, etc.). In particular, Deep Learning aims to \textit{learn} appropriate features directly from (often unstructured) data, and perform classification using these features. For Malicious URL Detection, this can help us reduce painful feature engineering, and build models without any domain expertise. With this motivation, in recent years, Deep Learning has been successfully applied for Malicious URL Detection \cite{saxe2017expose,le2018urlnet,benavides2019classification}. Among the early deep learning efforts, \cite{wang2016deep, yan2018detecting} used autoencoders to obtain a reduced dimension vector representation of the JavaScript code of a URL. This was followed by applying traditional classification models on the obtained representation. \cite{selvaganapathy2018deep} applied Deep Belief Networks to specific hand-designed features. \cite{saxe2018deep} use regular-expression based features extracted from the HTML-content and apply feedforward neural networks. Even Deep Reinforcement learning has been applied for malicious URL Detection \cite{chatterjee2019deep}

Due to the success of Bag-of-Words models in Malicious URL Detection, a natural extension to apply NLP-based deep learning models for this task, due to their ability to infer features from unstructured textual data. Consequently, Convolutional Neural Networks (CNNs) were applied for this task. eXpose \cite{saxe2017expose} applied character-level convolutional networks \cite{zhang2015character} to the URL string, with the hope of finding patterns of specific characters occurring together. This model did not use encoding for words, which meant that unlike the previous approaches which relied heavily on Bag-of-words, this model did not suffer from the problem of explosion in terms of number of features. Despite this, the model gave superior performance as compared to applying SVMs on Bag-of-Words features. URLNet \cite{le2018urlnet} developed an end to end system where convolutions were applied for both characters and words in the URL, and empirically demonstrated that using word-level information along with character-level could boost the performance of the model. To circumvent the problem of too many words, they developed a character-level word embedding, where the word embedding was based on the characters present in that word. 

There have been other parallel efforts using CNNs for some other variations for this problem (e.g \cite{jiang2017deep,abdi2017malicious}). \cite{shibahara2017malicious} applied denoising CNNs to detect anomalies in a session where compromised benign websites may redirect the users to malicious URLs. \cite{shima2018classification} tried to use advanced vector embeddings for characters, where they tried to encode neighboring character information as well. This was followed by the application of CNNs. \cite{yu2018character} used Deep Learning (LSTM \cite{hochreiter1997long} as well as CNN) for detecting algorithmically generated domain names.  There were other efforts combining CNNs with LSTMs \cite{pham2018exploring,yang2019phishing}. In a similar fashion, \cite{yang2019detecting} combined CNNs with a Gated Recurrent Unit to perform malicious URL detection. Some extensions of the LSTM-based models were investigated recently, where an attention mechanism was added to CNN and LSTM based models \cite{peng2019joint}, and Bidirectional LSTMs were applied \cite{wang2019bidirectional}. 

While delivering promising results, deep learning approaches are computationally very intensive, requiring GPUs for training. This problem gets amplified when new URLs (or data) becomes available, and the model needs to be retrained. Emerging approaches in the field of deep learning such as Lifelong Learning \cite{lopez2017gradient} and Online Deep Learning \cite{sahoo2018online} offer promising directions to address some of these challenges. 
\subsubsection{Feature Selection and Sparsity Regularization}
There are a large number and variety of features used for Malicious URL Detection, and there have been studies to determine which features maybe the most suitable \cite{buber2017feature,zuhair2016feature}. In particular, the usage of Bag of Words features for many of the feature categories results in millions of features. Moreover, as the number of URLs to be processed increases (which is the case in the real world setting), the number of features keeps growing as well. Learning prediction models using so many features suffers from two main drawbacks:
\begin{itemize}
	\item {Computationally Expensive}: Training and test time become significantly high, not only because of the many mathematical operations to be performed, but also because of collecting and preprocessing so many features. In many cases (e.g. using bi-gram and tri-gram features), we obtain so many features that it is practically infeasible to perform any optimization.
	\item {Noisy Models}: Malicious URL Detection often exhibits number of features being larger than the number of instances available for training. Optimizing such models may result in overfitting.
\end{itemize}

To overcome these problems, researchers in both machine learning and cyber-security have proposed representation learning techniques, which in the context of Malicious URL Detection are mostly concentrated in the domain of feature selection techniques, i.e., the optimal subset of the given representation needs to be learnt. Here we discuss two categories of representation learning: feature selection, where the features are evaluated and selected based on their performance, and sparsity regularization, where the feature selection is implicitly done by incorporating it into the objective function.

\textbf{\emph{Feature Selection}}.
There are two categories of feature selection where the features are scored on the basis of their performance and accordingly selection. These are  \emph{Filter Methods and Wrapper Methods} \cite{guyon2003introduction,zuhair2016feature}.

Filter methods usually use a statistical measure to evaluate the suitability of a particular feature. Based on this score, a set of features can be selected (and the poor features are filtered out). In most cases, this evaluation is done independently (i.e., independent of other features). Some popular approaches include the Chi squared test ($\chi^2$) \cite{pan2006anomaly,zhang2014domain}  and information gain scores \cite{kolari2006svms,ma2009detecting,toolan2010feature}.

Wrapper methods try to select the best subset of features, by modeling the feature selection as a search problem. Different subsets of features are used to learn a prediction model, and are evaluated based on their performance \cite{khonji2011study,basnet2012feature}. \cite{zhang2016application} use Genetic Algorithms to perform feature selection and divides the features into critical and non-critical. The critical features are used as is, while projection of the non-critical features are used to provide supplementary information, instead of being discarded. \cite{zuhair2016feature} adopt a maximum relevance and minimum redundancy criterion to select a robust subset of features. Other approaches used more recent sophisticated Grey Wolf Optimization based Rough Set Theory algorithm for feature selection \cite{rajitha2017suspicious}. \cite{chiew2019new} develop a 2-phase Hybrid Ensemble Feature Selection method, where first a cumulative distribution function gradient is exploited to produce primary feature subsets, which are fed into a data perturbation ensemble to obtain the secondary featrue subset. 

\textbf{\emph{Sparsity Regularization}}.
Due to a large number of features that are collected, in particular, lexical features, feature selection needs to be performed over millions of features. Applying filter and wrapper methods may not be practical. Feature selection can be induced by modifying the objective function, i.e., (called \emph{Embedded methods}) to embed the feature selection into the optimization. Consider the generic optimization problem discussed before:
\begin{eqnarray}
\nonumber
\min_\f \sum_{t=1}^T \ell(\f(\x_t), y_t) + \lambda\mathcal{R}(\mathbf{w})
\end{eqnarray}

where the first term optimizes model performance, and the second term is the regularizer. $\lambda$ is a tradeoff parameter to regulate the effect of regularization. A common technique to induce feature selection (or learn sparse models), is to use L1-norm regularizer:
\bq
\mathcal{R}(\mathbf{w}) = || \w ||_1
\eq
Unlike other popular regularizers (like L2-norm), which try to reduce model complexity, L1-regularization encourages zero values in the weight vector $\w$, which results in the corresponding features to not be selected for the final prediction model. This approach has been commonly used in conjunction with SVM and Logistic Regression models \cite{thomas2011design} in the batch setting. From the perspective of Online Setting, {Sparse Online Learning} \cite{wang2014high,wu2016sol} has been developed to learn sparse models for high-dimensional sparse data (as exhibited in URL features that are very sparse and in the order of millions).  \cite{hoi2012online,wang2014online} proposed \emph{Online Feature Selection} as a principled way for feature selection in an online learning manner, which aims to bound the number of selected features based on some given budget based on the first order online learning approaches. There is also a second-order Online Feature Selection approach \cite{wu2014massive}.

\subsection{Other Learning Methods}

While most of the algorithms used for Malicious URL Detection take the form of binary classification, other problem settings have also been studied. Some of these settings are designed for problems specifically arising in Malicious URL Detection. These include application of unsupervised learning to improve detection, learning similarity functions for detecting URLs, and learning interpretable models by string pattern mining for matching URLs. In the following we briefly discuss these problem settings for application to Malicious URL Detection.

\subsubsection{Unsupervised Learning}
Unsupervised learning is the scenario where the true label of the data is not available during the training phase. These approaches rely on anomaly detection techniques where the anomaly is defined as an abnormal behavior. There are several algorithms in literature that can be used for anomaly detection (e.g. clustering, 1-class SVMs, etc.). Unfortunately, due to the extremely diverse set of URLs, it is hard to determine what is "normal" behavior and what is an anomaly. As a result, such techniques have not become very popular for malicious URL detection. However, some techniques have tried to use unsupervised techniques in conjunction with supervised techniques to improve performance. For example, \cite{yoo2014two} integrate supervised and unsupervised learning in two stages. In the first stage, the supervised prediction model predicts malicious URLs. For those that are classified as benign, a 1-class SVM is used to look for anomalies. \cite{feroz2015phishing} follow another style of integration of supervised and unsupervised where first k-means clustering is performed, and the cluster ID is used as a feature to train a classifier. \cite{popescu2015study} proposed the usage of an unsupervised hash-based clustering system, wherein specific clusters were labeled as malicious or benign (based on majority of the training data).

\subsubsection{Similarity Learning}
Similarity learning aims to learn how similar two instances are (or in our case, how similar 2 URLs are). This field helps us identify which specific legitimate URLs are being mimicked by the attackers. For this setting, there is a set of protected URLs, and a set of suspicious URLs which are potentially trying to mimic the protected URLs. The aim is to measure the similarity of the suspicions URLs with the protected URLs, and if the similarity is above a specific threshold, we are able to spot a malicious URL. This setting has been largely addressed by extracting visual features \cite{liu2006antiphishing,wenyin2005detection,chen2009fighting,afroz2011phishzoo}, and computing the similarity between the images of the suspicious and the protected URLs.

Another related area is learning using kernel functions, which are notions of similarity, and also allow models to learn nonlinear classifiers \cite{Smola1998}. The prediction function takes the following form:
\begin{eqnarray}
\f_t(\x_t) = \sum_{i=1}^{t-1}\alpha_i\kappa(\x_i, \x_t)
\end{eqnarray}
where $\alpha_i$ is the coefficient for each instance learnt by some learning algorithm, and $\kappa$ denotes a kernel function (e.g., a polynomial or Gaussian kernel) which measures the similarity between the two instances. Whenever $\alpha_i\neq 0$, the training instance is often known as a support vector. There has been abundant literature on using kernels both in batch and online settings. The batch setting requires a lot of memory and computational resources, which is often intractable in the real world. Online Learning with kernels \cite{kivinen2004online} follows similar approaches as most other online learning techniques except that the pool of support vectors often will increasingly expand whenever there is a mistake in the online learning process. This will result in a very undesired drawback, i.e., the unbounded support vector size, which not only increases the computational cost but also the need for huge memory space for storing all the support vectors in online learning systems. To address this problem, budget online kernel learning has been extensively studied in online learning. Examples include Randomized Budget Perceptron \cite{cavallanti2007tracking}, Forgetron \cite{Dekel2008}, Projectron \cite{Orabona2008}, and Bounded Online Gradient Descent \cite{zhao2012fast}. Some of these techniques were empirically studied by \cite{ma2009identifying} for malicious URL detection, but they did not manage to get satisfactory performance. This is not a surprising result, as any small budget size that would enable scalable computation would miss out on most of the instances to be stored as URLs, and will give a very poor kernel approximation. With the recent development of efficient functional approximation techniques for large-scale online learning with kernels \cite{wang2013large,lu2016large}, it may be possible to obtain competitive performance. Further, Online Multiple Kernel Learning \cite{hoi2013online,sahoo2014online} approaches allow for learning with multi-modality, wherein different feature sets correspond to different modalities.

\subsubsection{String Pattern Matching}
As discussed in the previous sections, lexical features are often obtained in the form of bag-of-words representation, which is often of abnormally high dimensionality and could grow over time. Using such predefined features may not be practical for real-world deployment, and such an approach may cause hindrance to interpreting particular common types of attacks. Further, such techniques cannot identify signatures in the form of substrings. To address these issues, \cite{huang2014malicious} proposed a dynamic string pattern mining approach which borrowed the ideas from efficient searching and substring matching. A similar approach based on Trigrams was also developed by \cite{xiong2015mird}. \cite{prakash2010phishnet,sun2015autoblg,bo2016automating} also designed an approximate string matching strategy to improve over the exact match required by blacklists. While these methods performed string pattern analysis on the URL string, \cite{choi2010automatic} performed string pattern analysis on JavaScript features.

\subsection{Summary of Machine Learning Algorithms}

There are a a wide variety of machine learning algorithms in literature that can be directly used in the context of Malicious URL Detection. Due to potentially a tremendous size of training data (millions of instances and features), there was a need for scalable algorithms, and that is why Online Learning methods have found a lot of success in this domain. Efforts have also been made to automatically learn the features, and to perform feature selection. Lastly, there have been efforts in modifying the problem from a typical binary classification algorithm to address class imbalance and multi-class problems. We categorize the representative references according to the machine learning methods applied in Table \ref{tab:ml}. Using these technologies to build live systems is another challenging task. In the following we discuss real systems to demonstrate how Malicious URL Detection can be used as a service. 

\begin{table*}[htbp]
	\small
	\centering
	\caption{Representative References of different types of Machine Learning Algorithms used for Malicious URL Detection}
	\begin{tabular}{|c|c|c|}
		\hline
		{\textbf{Methodology}} & \textbf{Sub Category} & {\textbf{Representative References}} \\
		\hline
		\hline
		\multirow{7}[0]{*}{\textbf{Batch Learning}} & \multirow{2}[0]{*}{{SVM}}  &  \cite{kolari2006svms,pan2006anomaly,ma2009beyond,hou2010malicious,bannur2011judging,he2011efficient,huang2012svm,pao2012malicious,chiba2012detecting}\\
		&& \cite{chu2013protect,sorio2013detection,wang2013click,xu2013cross,marchal2014phishscore,marchal2014phishstorm,alshboul2015detecting,jain2017towards,verma2017s,wu2018malicious,altay2018context}\\
		\cline{2-3}
		& Logistic Regression &  \cite{garera2007framework,ma2009beyond,bannur2011judging,canali2011prophiler,lee2012warningbird,wang2013click,xu2013cross,jain2017towards,jain2019machine}\\
		\cline{2-3}
		& Naïve Bayes & \cite{hou2010malicious,canali2011prophiler,aggarwal2012phishari,xu2013cross,cao2016detection,jain2017towards,wang2017tsmwd} \\
		\cline{2-3}
		&\multirow{1}[0]{*}{{Decision Trees}} & \cite{seifert2008identification,bilge2011exposure,canali2011prophiler,aggarwal2012phishari,wang2013click,xu2013cross,cao2016detection,marchal2014phishscore}\\
		&& \cite{marchal2014phishstorm,alshboul2015detecting,jain2017towards,buber2017nlp,patil2018malicious, sahingoz2019machine,cuzzocrea2019machine}\\
		\cline{2-3}
		& Others and Ensembles& \cite{astorino2016malicious,choi2011detecting,eshete2013binspect,su2013suspicious,eshete2013einspect,eshete2013effective} \\
		\hline
		\multirow{4}[0]{*}{\textbf{Online Learning}} & First-order algorithms & \cite{Rosenblatt1958,Zinkevich2003,Crammer2006,ma2009identifying,ma2011learning,thomas2011design,zhang2011malicious,verma2017s} \\
		\cline{2-3}
		& Second-order algorithms &  \cite{dredze2008confidence,ma2009identifying,blum2010lexical,ma2010exploiting,crammer2009adaptive,le2011phishdef,lin2013malicious,verma2017s}\\
		\cline{2-3}
		& Cost-Sensitive Online Learning &  \cite{WZH14,zhao2013cost}\\
		\cline{2-3}
		& Online Active learning & \cite{sculley2007online,zhao2013cost,lu2016online,bhattacharjee2017prioritized} \\
		\hline
		\multirow{4}[0]{*}{\textbf{Representation Learning}} & \multirow{2}[0]{*}{ Deep Learning} &  \cite{wang2016deep,jiang2017deep,shibahara2017malicious,abdi2017malicious,saxe2017expose,selvaganapathy2018deep,yan2018detecting,saxe2018deep}\\
		&& \cite{shima2018classification,yu2018character, pham2018exploring,yang2019detecting,peng2019joint, wanda2019urldeep, chatterjee2019deep, li2019domain,yang2019phishing}\\
		\cline{2-3}           
		& \multirow{1}[0]{*}{{Feature Selection}} & \cite{guyon2003introduction,zuhair2016feature,pan2006anomaly,zhang2014domain,kolari2006svms,ma2009detecting,toolan2010feature,khonji2011study,basnet2012feature,zhang2016application,chiew2019new} \\
		\cline{2-3}
		& Sparsity Regularization & \cite{thomas2011design,wang2014high,wu2016sol,hoi2012online,wang2014online,wu2014massive}\\
		\hline
		\multirow{3}[0]{*}{\textbf{Other Learning}}
		
		& Unsupervised Learning  & \cite{yoo2014two,feroz2015phishing,popescu2015study} \\
		\cline{2-3}
		& Similarity Learning & \cite{liu2006antiphishing,wenyin2005detection,chen2009fighting,afroz2011phishzoo} \\
		\cline{2-3}
		& String Pattern Matching & \cite{wardman2009identifying,huang2014malicious,xiong2015mird,prakash2010phishnet,sun2015autoblg,bo2016automating,choi2010automatic} \\
		\hline
	\end{tabular}%
	\label{tab:ml}%
\end{table*}%

\section{Malicious URL Detection as a Service}

Malicious URL detection using machine learning can have immense real-world applications. However, it is nontrivial to make it work in practice.
Several researchers have proposed architectures for building end-to-end malicious URL Detection systems and deployed them for real-world utilities. Many have focused on providing services to Online Social Networks like Twitter, where users share plenty of URLs (\cite{thomas2011design,lee2012warningbird,cao2015detecting,alshboul2015detecting} etc.)

\subsection{Design Principles}
When designing and building a real-world malicious URL detection system using machine learning techniques, we aim to achieve several goals. There are many objectives and parameters that are needed to traded-off in order to achieve the desired results. We briefly discuss the most important factors for consideration below:

(i) \emph{Accuracy}: This is often one of the most important goals to be achieved for any malicious URL detection. Ideally, we want to maximize the detection of all the threats of malicious URLs (``true positives") while minimizing the wrong detection of classifying benign URLs into malicious (``false positives"). Since no system is able to guarantee a perfect detection accuracy, a practical malicious URL detection system often has to trade off between the ratios of false positives and false negatives by setting different levels of detection thresholds according to the application needs (e.g. security concerns vs. user experience).


(ii) \emph{Speed of Detection}: Detection speed is an important concern for a practical malicious URL detection system, particularly for online systems or cybersecurity applications. For example, when deploying  the malicious URL detection service in online social networks like Twitter, whenever a user posts any new URL, an ideal system should be able to detect the malicious URL immediately and then block the URL and its related tweets in real time to prevent any threats and harms to public. For some cybersecurity applications, the requirement of detection speed could be more severe, which sometimes needs the detection to be done in milliseconds such that a malicious URL request can be blocked immediately in real time whenever upon a user click.



(iii) \emph{Scalability}: Considering the increasing huge amount of URLs, a real-world malicious URL detection system must be able to scale up for training the models with millions or even billions of training data. In order to achieve the high scalability desire, there are two major kinds of efforts and solutions. The first is to explore more efficient and scalable learning algorithms, e.g., online learning or efficient stochastic optimization algorithms. The second is to build scalable learning systems in distributed computing environments, e.g., using emerging distributed learning frameworks (such as Apache Hadoop or Spark) on the clusters.


(iv) \emph{Adaptation}:
A real-world malicious URL detection system has to deal with a variety of practical complexities, including adversarial patterns such as concept drifting where the distribution of malicious URLs changes over time or even change in adversarial way to bypass the detection system, missing values (e.g., unavailable features or too expensive to compute under given time constraints), increasing number of new features, etc. A real-world malicious URL detection system must have a strong adaptation ability to work effectively and robustly under most circumstances.


(v) \emph{Flexibility}:
Given the high complexity of malicious URL detection, a real-world malicious URL detection system with machine learning should be designed with good flexibility for easy improvements and extensions. These include the quick update of the predictive models with respect to new training data, being easy to replace the training algorithm and models whenever there is a need, being flexible to be extended for training models to deal with a variety of new attacks and threats, and finally being able to interact with human beings whenever there is a need, e.g., active learning or crowdsourcing for enhancing performance.

\subsection{Design Frameworks}

In the following, we discuss some real-world implementations, heuristics and systems that have attempted to make Malicious URL Detection as a service in reality.

\cite{thomas2011design} designed a framework called \emph{Monarch}, and the idea of providing Malicious URL Detection as a service was floated. Monarch would crawl web-services in real-time and determine whether the URL would be malicious or benign. Differences in malicious URL distribution between Twitter and Spam Emails were observed, and accordingly different models were built for both. Monarch, at the time of development, could process 15-million URLs a day for less than \$800 per day. The implementation of this system comprises a URL aggregator which collects URLs from some data streams (e.g., Twitter or Emails). From these URLs, features are collected, which are then processed and converted into sparse features in the feature extractor. Finally, a classifier is trained on the processed data to detect malicious URLs. The collected features include both URL-based features and content-based features. In addition, initial URLs, landing URLs, and Redirects are also extracted. For fast classification training from the perspective of efficiency in memory and algorithmic updates, a linear classifier based on logistic regression with the L1-regularizer (for inducing sparse models) is trained on a distributed computing architecture \cite{singer2009efficient,mcdonald2010distributed}. They are able to process an entire URL in 5.5 seconds on average, most of which is spent on feature extraction. The prediction is relatively efficient.

\cite{whittaker2010large} applied machine learning techniques to predict malicious URLs, and subsequently attempted to maintain a blacklist of malicious URLs using these predictions. \emph{Prophiler} \cite{canali2011prophiler} is another system, where it recommends that the URL classification to be performed in a two-stage process. The first stage would be by analyzing the light-weight URL features, and quickly filtering out URLs for which the classifier has a high confidence. For the low confidence predictions, a more intensive content based analysis can be performed. \emph{WarningBird} \cite{lee2012warningbird} is similar to Monarch in that the primary aim is to detect suspicious URLs in a Twitter streams except that it uses heuristics to obtain new features, and the classifier used was an SVM model using LIBLINEAR \cite{fan2008liblinear} and not trained on a distributed architecture. BINSPECT \cite{eshete2013binspect} is another system that was developed to take advantage of ensemble classification, where the final prediction was made on the basis of confidence weighted majority voting.

\section{Practical Issues and Open Problems}

Despite many exciting advances over last decade for malicious URL detection using machine learning techniques, there are still many open problems and challenges which are critical and imperative, including but not limited to the following:


(i) \emph{High volume and high velocity}: Real-world URL data is a form of big data with high volume and high velocity. In August 2012 \cite{sullivan2012google} disclosed that Google's search engine found more than 30 trillion unique URLs on the Web, and crawls 20 billion sites a day. It is almost impossible to train a malicious URL detection model on all the URL data. A clever way of sampling effective URL training data (including both malicious and benign ones) and training them using efficient and scalable machine learning algorithms will be always an open question for researchers in both machine learning and cybersecurity communities.


(ii) \emph{Difficulty in acquiring labels}: Most existing malicious URL detection approaches by machine learning are based on supervised learning techniques, which require labeled training data including both benign and malicious URL data. The labeled data can be obtained by either asking human experts to label or acquiring from blacklists/whitelists (which were often also labeled manually). Unfortunately the scale of such labeled data is tiny as compared to the size of all available URLs on the web. For example, one of the largest public available malicious URL training data sets in academia \cite{ma2009identifying} has only 2.4 million URLs. Thus, there is a large room for open research in resolving the difficulty of acquiring labeled data or learning with limited amount of labeled data. One possible direction is to go beyond purely supervised learning by exploring unsupervised learning or semi-supervised learning, such as active learning in some recent studies \cite{zhao2013cost}. Another possible direction is to explore crowdsourcing techniques \cite{hu2011poster,doan2011crowdsourcing} by facilitating organizations and individuals to label and share malicious URLs. However this is not trivial given the practical concerns of cost, privacy and security threats of sharing malicious URLs. More innovations could be explored in the future.


(iii) \emph{Difficulty in collecting features}: As mentioned in the previous section, collecting features for representing a URL is crucial for applying machine learning techniques. However, it is often not a trivial task for collecting many kinds of features for a URL. In particular, some features could be costly (in terms of time) to collect, e.g., host-based features. Some features might be missing, or noisy, or can not be obtained due to a variety of reasons (e.g., IP/DNS addresses of a URL may vary time to time). In addition, real-world URLs may not always be alive. For example, as observed by \cite{mcgrath2008behind}, many malicious URLs may be short lived, and thus accessing its features (e.g., IP address) may not be possible when it is not alive. Besides, some previous benign URLs may be stopped for services, and then were replaced by some malicious URL (or vice versa). All these challenges pose a lot of research and development difficulties for collecting features (especially for constructing training datasets).


(iv) \emph{Feature Representation}: In addition to high volume and high velocity of URL data, another key challenge is the very high-dimensional features (often in millions or even billion scale). This poses a challenge in practice for training a classification model. Some commonly used learning techniques, such as feature , dimension reduction and sparse learning, have been explored, but they are far from solving the challenge effectively. Besides the high dimensionality issue, another more severe challenge is the evolving high dimensional feature space, where the feature space often grows over time when new URLs and new features are added into the training data. This again poses a great challenge for a clever design of new machine learning algorithms which can adapt to the dynamically changing feature spaces. Recently \cite{saxe2017expose,le2018urlnet} have developed promising solutions to learn the feature representation through deep learning. There have also been approaches to use Convolution networks for transfer learning for malicious URL detection \cite{rajalakshmi2018transfer}.


(v) \emph{Concept drifting and emerging challenges}: Another challenge is concept drifting where the distribution of malicious URLs may change over time due to the evolving behaviour of new threats and attacks. This requires machine learning techniques to deal with concept drifting whenever it appears \cite{tan2018adaptive,arnaldo2018acquire}. Besides, another recent challenge is due to the popularity of URL shortening services, which take a long URL as input and produce a short URL as an output. Such URL shortening services potentially offer an excellent way for malicious hackers and criminals to hide their malicious URLs and thus creates a new challenge for automated malicious URL detection systems. Lastly, it is almost certain that there will always new types of challenges for malicious URL detection since sophisticated malicious hackers and criminals will always find ways to bypass the cyber security systems. How to make effective learning systems which can quickly detect and adapt themselves for resolving new challenges will be a long term research challenge. An additional issue is to identify vulnerable websites which may become malicious in the future \cite{soska2014automatically}. This is important, as URLs deemed to be benign in the past may get compromised and become malicious in the future.

(vi) \emph{Interpretability of Models}: An important research direction is to understand what makes a URL malicious, and to validate what patterns in URLs can help determine the malicious or benign nature of a URL. This is particularly difficult when deep learning models are used, which often behave like black boxes. Some visualizations of the URL embeddings were shown to identify specific string patterns in URLs in \cite{le2018urlnet}. For example, they observed that when "/opt/<PATH>" appeared in the URL, where the path could be any path, the URL was likely to be malicious. \cite{pang2019deep} solved malicious URL detection as a deep anomaly detection task, and showed how to interpret the anomaly scores. MADE \cite{oprea2018made}, a recent enterprise threat detection system was designed to provide interpretable results. Obtaining a deeper understanding of malicious URLs, specifically why they are malicious can have immense applications in designing modern security systems. 

(vii) \emph{Adversarial Attacks}: As the machine learning models become better at detecting malicious URLs, it is only natural that attackers will try to adversarially react to improve the quality of attacks. Recent efforts have studied how machine learning models can simulate URLs to bypass machine learning based malicious URL detectors \cite{bahnsen2018deepphish}. In fact some recent approaches have used the adversarial loss principle \cite{goodfellow2014generative} for training malicious URL detectors \cite{anand2018phishing,trevisan2019robust}. A systematic study of how the trained models would behave in an adversarial environment can thus help identify the limitations of the current models \cite{shirazi2019adversarial}, and provide directions for future research.


\section{Related Surveys}

Recent years have witnessed innovative applications of machine learning in cyber security \cite{singh2013survey,dua2016data,buczak2016survey}. They discuss a variety of other cyber threats, and do not focus on Malicious URL Detection. For example, \cite{buczak2016survey} present a survey on the usage of machine learning and data mining techniques for Cyber Security intrusion detection. While there are surveys on Malicious URL Detection using Machine Learning, most are either limited in scope or outdated. For example \cite{abu2007comparison} did an empirical analysis of different machine learning techniques for Malicious URL Detection in 2007, at a time when neither features nor machine learning models for this task had been extensively explored.
\cite{khonji2013phishing,patil2015survey} gave a broad overview of Phishing and its problems, but do not extensively survey the feature representation or the learning algorithms aspect.
\cite{zuhair2016feature} focused on primarily feature selection for Malicious URL Detection.

Malicious URL Detection is closely related to other areas of research such as Spam Detection. \cite{spirin2012survey} conducted a comprehensive survey in 2012, wherein they identified different types of spam(Content Spam, Link Spam, Cloaking and Redirection, and Click Spam), and the techniques used to counter them. They categorized the Spam Detection techniques into Content-based Spam Detection (using lexical features such as Bag of Words and Natural Language Processing techniques), Link-based spam detection (utilizing the information regarding the connectivity of different URLs) and other miscellaneous techniques. Spam Detection is heavily reliant on processing the text in an email and utilizing natural language processing for analysis. These techniques are not directly useful for Malicious URL Detection, unless they are used to draw inference about the context in which the URL has appeared. Despite some overlap between the techniques used for spam detection and malicious URL detection, spam detection techniques largely qualify as techniques that use context-based features for detecting malicious URLs. Other recent learning based spam detection surveys include \cite{blanzieri2008survey,crawford2015survey,heydari2015detection}, many of which focus on spam appearing in online reviews.

Another closely related area is Webpage Classification. \cite{qi2009web} conducted a survey on the features and algorithms deployed for webpage classification. The most common types of features used are the content features (text and HTML tags on the page), and Features of Neighbors (classification based on the the class label of similar webpages). After the feature construction, standard classification techniques are applied, often with focus on multi-class classification and hierarchical classification. Like Spam detection, webpage classification also benefits significantly from text classification techniques.

Spam Detection, Webpage Classification, and Malicious URL Detection, all use a few similar types of features  and techniques to solve the problem. In practice, these methods and features can complement each other to improve the performance of the machine learning models. In general, the features used for Spam Detection and Web-page classification are a subset of those features that are commonly used for Malicious URL Detection.

\section{Concluding Remarks}

Malicious URL detection plays a critical role for many cybersecurity applications, and clearly machine learning approaches are a promising direction. In this article, we conducted a comprehensive and systematic survey on Malicious URL Detection using machine learning techniques. In particular, we offered a systematic formulation of Malicious URL detection from a machine learning perspective, and then detailed the discussions of existing studies for malicious URL detection, particularly in the forms of developing new feature representations, and designing new learning algorithms for resolving the malicious URL detection tasks. In this survey, we categorized most, if not all, the existing contributions for malicious URL detection in literature, and also identified the requirements and challenges for developing Malicious URL Detection as a service for real-world cybersecurity applications.

Finally, we highlighted some practical issues for the application domain and indicated some important open problems for further research investigation. In particular, despite the extensive studies and the tremendous progress achieved in the past few years, automated detection of malicious URLs using machine learning remains a very challenging open problem. Future directions include more effective feature extraction and representation learning (e.g., via deep learning approaches), more effective machine learning algorithms for training the predictive models particularly for dealing with concept drifts (e.g., more effective online learning) and other emerging challenges (e.g., domain adaption when applying a model to a new domain), and finally a smart design of closed-loop system of acquiring labeled data and user feedback (e.g., integrating an online active learning approach in a real system).


\bibliographystyle{ACM-Reference-Format}
\bibliography{mud_bib}

\end{document}